\documentclass[]{gcs}

\RequirePackage[
  datamodel=acmdatamodel,
  style=acmauthoryear,
  backend=biber,
  giveninits=true,
  uniquename=init
]{biblatex}

\usepackage{xcolor}

\usepackage{tikz}
\usepackage{subcaption}
\usepackage{amsmath}%
\usepackage{amsthm}%
\usepackage{amsfonts}%
\usepackage{mathrsfs}%
\usepackage{mathtools}%

\usepackage[linesnumbered, title numbered, ruled, noend]{algorithm2e}
\SetAlFnt{\small}

\newtheorem{example}{Example}
\newtheorem{theorem}{Theorem}
\newtheorem{definition}{Definition}

\usepackage{etoolbox}
\usepackage{xspace}

\newbool{CPOptimizerCited}
\newcommand{\CPOptimizer}{%
  CPOptimizer%
  \ifbool{CPOptimizerCited}{}{\nobreakspace\cite{laborie2018ibm}\global\booltrue{CPOptimizerCited}}%
  \xspace}

\newbool{OptalCPCited}
\newcommand{\OptalCP}{%
  OptalCP%
  \ifbool{OptalCPCited}{}{\nobreakspace\cite{OptalCP}\global\booltrue{OptalCPCited}}%
  \xspace}

\newbool{MaxiCPCited}
\newcommand{\MaxiCP}{%
  MaxiCP%
  \ifbool{MaxiCPCited}{}{\nobreakspace\cite{maxicp}\global\booltrue{MaxiCPCited}}%
  \xspace}

\newbool{GecodeCited}
\newcommand{\Gecode}{%
  Gecode%
  \ifbool{GecodeCited}{}{\nobreakspace\cite{schulte2006gecode}\global\booltrue{GecodeCited}}%
  \xspace}

\newbool{MiniZincCited}
\newcommand{\MiniZinc}{%
  MiniZinc%
  \ifbool{MiniZincCited}{}{\nobreakspace\cite{optionminizinc}\global\booltrue{MiniZincCited}}%
  \xspace}

\newbool{MiniCPCited}
\newcommand{\MiniCP}{%
  MiniCP%
  \ifbool{MiniCPCited}{}{\nobreakspace\cite{MiniCP}\global\booltrue{MiniCPCited}}%
  \xspace}

\newbool{SICStudCited}
\newcommand{\SICStus}{%
  SICStus%
  \ifbool{SICStudCited}{}{\nobreakspace\cite{sicstusmanual}\global\booltrue{SICStudCited}}%
  \xspace}

\newcommand{\gcumulative}{\texttt{GeneralizedCumulative}}

\newcommand{\minC}{\underline{\mathcal{C}}}
\newcommand{\maxC}{\overline{\mathcal{C}}}

\newcommand{\minP}{\underline{P}}
\newcommand{\maxP}{\overline{P}}

\newcommand{\mins}{\underline{s}}
\newcommand{\maxs}{\overline{s}}

\newcommand{\mine}{\underline{e}}
\newcommand{\maxe}{\overline{e}}

\newcommand{\mind}{\underline{d}}
\newcommand{\maxd}{\overline{d}}

\newcommand{\minc}{\underline{c}}
\newcommand{\maxc}{\overline{c}}

\begin{document}

\title[Generalized Cumulative]{Implementing Cumulative Functions with Generalized Cumulative Constraints}


\author{Pierre Schaus}
\authornote{Corresponding Author.}
\orcid{0000-0002-3153-8941}
\email{pierre.schaus@uclouvain.be}
\affiliation{%
  \institution{UCLouvain, ICTEAM}
  \city{Louvain-la-Neuve}
  \country{Belgium}
}

\author{Charles Thomas}
\orcid{0000-0002-7360-5372}
\email{cftmthomas@gmail.com}
\affiliation{%
  \institution{UCLouvain, ICTEAM}
  \city{Louvain-la-Neuve}
  \country{Belgium}
}

\author{Roger Kameugne}
\orcid{0000-0003-1809-9822}
\email{rkameugne@gmail.com}
\affiliation{%
  \institution{UCLouvain, ICTEAM and University of Maroua}
  \city{Louvain-la-Neuve / Maroua}
  \country{Belgium / Cameroon}
}

\renewcommand{\shortauthors}{Schaus, Thomas, \& Kameugne}

\begin{abstract}
Modeling scheduling problems with conditional time intervals and cumulative functions has become a common approach when using modern commercial constraint programming solvers.
This paradigm enables the modeling of a wide range of scheduling problems, including those involving producers and consumers. 
However, it is unavailable in existing open-source solvers and practical implementation details remain undocumented.
In this work, we present an implementation of this modeling approach using a single, generic global constraint called the Generalized Cumulative. 
We also introduce a novel timetabling filtering algorithm specifically designed to handle tasks defined on conditional time-intervals. 
Experimental results demonstrate that this approach, combined with the new filtering algorithm, performs competitively with existing solvers enabling the modeling of producer and consumer scheduling problems and effectively scales to large-scale problems.
\end{abstract}




\maketitle

\section{Introduction}

Constraint Programming (CP) is a declarative paradigm for solving
combinatorial satisfaction and optimisation problems.
A CP model consists of a set of \emph{decision variables},
each associated with a finite domain of possible values,
and a collection of \emph{constraints} that restrict the combinations of values that the variables may simultaneously take.
Solving alternates between \emph{constraint propagation} --- removing domain values that cannot appear in any solution --- and \emph{search}, typically a form of systematic backtracking. A key strength of CP is the availability of \emph{global constraints}: high-level constraints such as \texttt{AllDifferent}~\cite{regin1994filtering}
or \texttt{Regular}~\cite{pesant2004regular} that capture structured combinatorial patterns and come with dedicated, efficient propagation algorithms (see~\cite{catalogue} for an extensive catalogue). We refer the reader to~\cite{MiniCP} for a comprehensive introduction to CP and CP solvers.

CP is particularly well-suited to scheduling, where the goal is to assign
integer-valued start times, durations, and resource consumptions to a set of tasks subject to temporal and capacity restrictions~\cite{Baptiste01,laborie2018ibm}.
The classical \textit{cumulative} constraint~\cite{aggoun1993extending}, for instance, takes a set of tasks each described by integer variables $(s_i, d_i, c_i)$ for start time, duration, and resource consumption, and enforces that at every time point the
total consumption of overlapping tasks does not exceed a fixed capacity.
Similarly, the \textit{disjunctive} constraint enforces that no two tasks overlap in time.
These constraints, together with efficient filtering algorithms, form the backbone of CP-based scheduling solvers~\cite{Baptiste01}.

The modeling flexibility of CP for scheduling has been further enhanced with the introduction of \emph{conditional time-interval variables}~\cite{LaborieR08}, which represent the execution of tasks that may or may not take place. 
Such tasks are called optional tasks.
These are convenient for modeling problems with alternative resources~\cite{kinable2014concrete,kumar2018station,cappart2017rescheduling,kizilay2018constraint}.
Building on top of conditional time-intervals, the same authors introduced an algebraic model for cumulative resources called cumulative functions~\cite{LaborieRSV09}. It provides an efficient and convenient way to model producer-consumer problems with optional tasks~\cite{liu2018modelling,gedik2018constraint,cappart2018constraint,thomas2024constraint}.
While commercial solvers such as \CPOptimizer and \OptalCP natively support conditional time-intervals and cumulative functions, both are closed-source engines whose concrete propagation algorithms and data structures are proprietary.
In the open-source domain, \Gecode offers the generalized cumulative constraint (\texttt{cumulatives}) of~\textcite{BeldiceanuC02}.
However, that constraint operates on flat arrays of integer variables rather than cumulative function expressions, encodes optionality indirectly through dummy resources, and relies on an event-sweep algorithm that exhibits several pruning limitations in practice.
Conversely, profile-based timetabling algorithms~\cite{GayHS15,GingrasQ16} were developed strictly for the classical cumulative constraint with mandatory tasks and positive consumptions.

In this paper, we bridge this gap:
(1) we provide an automated translation that flattens bound restrictions on cumulative functions into a generalized cumulative constraint;
(2) we generalize profile-based timetabling to simultaneously handle conditional time-intervals, variable and negative consumptions (producers and consumers), and variable durations, achieving strictly stronger pruning than the sweep algorithm of~\textcite{BeldiceanuC02}; and
(3) we contribute the first fully documented, open-source reference implementation of cumulative functions and generalized timetabling within the \MaxiCP solver.
The experimental results demonstrate that our algorithm achieves competitive filtering and scales effectively to large-scale scheduling problems.

The remainder of this article is organized as follows.
Section~\ref{sec:modeling} introduces the modeling background on conditional time-interval variables, cumulative functions, and the \gcumulative\ constraint.
Section~\ref{sec:timetabling} presents our timetabling filtering algorithm, including timeline initialization, the filtering rules, and its theoretical complexity.
Section~\ref{sec:rlw} reviews related work and contrasts our approach with existing cumulative algorithms.
Section~\ref{sec:xp} evaluates the performance of the algorithm on multiple benchmark scheduling problems across different solvers.
Finally, Section~\ref{sec:concl} concludes and discusses avenues for future work.

\section{Modeling with Cumulative Functions}\label{sec:modeling}

This section introduces the modeling primitives used throughout the paper.
We first define conditional time-interval variables and the elementary cumulative functions that describe a task's contribution to a resource profile.
Next, we show how to combine these functions to model complex resource profiles and flatten them into a set of cumulative tasks.
Finally, we formally define the \gcumulative\ constraint and explain how restrictions on cumulative functions are enforced through it.

\subsection{Conditional Time-Intervals}

Following the original terminology introduced by~\textcite{LaborieR08}, a conditional time-interval variable is also called an \emph{interval} for short. 

\begin{definition}[Conditional Time-Interval Variable]
\label{def:cti}
A \emph{conditional time-interval variable} $x$, introduced by~\textcite{LaborieR08}, is a variable whose domain is a subset of $\{\bot\} \cup \{[s,e) \mid s \leq e,\, s,e \in \mathbb{Z}\}$.
The value $x = \bot$ denotes that the interval is absent; the value $x = [s,e)$ denotes that it executes from $s$ to $e$.
Its domain is represented by a tuple $\langle s, d, e, p \rangle$ where:
\begin{itemize}
    \item $p \in \{\mathit{true}, \mathit{false}\}$ is the execution status ($\mathit{false}$ corresponds to $\bot$);
    \item $s \in [\mins, \maxs]$ and $e \in [\mine, \maxe]$ are the (non-negative) start and end time variables;
    \item $d \in [\mind, \maxd]$ is the (non-negative) duration variable.
\end{itemize}
Bound-consistent filtering on $s + d = e$ is maintained atomically: if propagation reduces any of the ranges $[\mins,\maxs]$, $[\mind,\maxd]$, $[\mine,\maxe]$ to empty (i.e.\ $\mins>\maxs$, $\mind>\maxd$, or $\mine>\maxe$), the interval is forced absent ($p \gets \mathit{false}$) if it is optional, or a contradiction is raised if $p = \mathit{true}$.
When $p = \mathit{false}$, the interval is absent and the attributes $s$, $d$, $e$ are irrelevant: the interval contributes nothing to any constraint.
We denote by $I$ the set of all such intervals in the problem. For interval~$i \in I$, its attributes are written $s_i, d_i, e_i, p_i$ with bounds~$\mins_i, \maxs_i$, $\mind_i, \maxd_i$, $\mine_i, \maxe_i$.
\end{definition}

\subsection{Cumulative Functions}
For modeling cumulative problems, \textcite{LaborieRSV09} introduced the idea of cumulative function expressions.
They allow the expression of step-wise integer functions of conditional time-interval variables with consumption on a resource.
The contribution of a single time-interval variable~$x\in I$ with a non-negative consumption $c \in \mathbb{Z}^+$ is represented by an elementary cumulative function. The elementary cumulative functions are \emph{pulse}, \emph{stepAtStart}, and \emph{stepAtEnd}~\cite{LaborieRSV09} as illustrated in Figure~\ref{fig:elementary}. 
\begin{figure}[tbp]

\centering


    \begin{subfigure}{0.32\textwidth}
        \centering
        \begin{tikzpicture}[scale=0.75]
            \draw[->] (-0.5,0) -- (4.5,0) node[right] {$t$};
            \draw[->] (0,-1.5) -- (0,3) node[above] {};

            \filldraw[fill=blue!40!white, draw=black] (1,-1) rectangle (3,-0.2);
            \node at (2,-0.6) {$x$};

            \draw[dotted] (0,2) node[left] {$c$} -- (1,2);
            \draw[thick, blue!80!black] (-0.5,0) -- (1,0) -- (1,2) -- (3,2) -- (3,0) -- (4.2,0);
        \end{tikzpicture}
        \caption{$\operatorname{pulse}(x, c)$}
    \end{subfigure}\hfill
    \begin{subfigure}{0.32\textwidth}
        \centering
        \begin{tikzpicture}[scale=0.75]
            \draw[->] (-0.5,0) -- (4.5,0) node[right] {$t$};
            \draw[->] (0,-1.5) -- (0,3) node[above] {};

            \filldraw[fill=blue!40!white, draw=black] (1,-1) rectangle (3,-0.2);
            \node at (2,-0.6) {$x$};
            \draw (3, 0.15) -- (3, -0.15); 

            \draw[dotted] (0,2) node[left] {$c$} -- (1,2);
            \draw[thick, blue!80!black] (-0.5,0) -- (1,0) -- (1,2) -- (4.2,2);
        \end{tikzpicture}
        \caption{$\operatorname{stepAtStart}(x, c)$}
    \end{subfigure}\hfill
    \begin{subfigure}{0.32\textwidth}
        \centering
        \begin{tikzpicture}[scale=0.75]
            \draw[->] (-0.5,0) -- (4.5,0) node[right] {$t$};
            \draw[->] (0,-1.5) -- (0,3) node[above] {};

            \filldraw[fill=blue!40!white, draw=black] (1,-1) rectangle (3,-0.2);
            \node at (2,-0.6) {$x$};
            \draw (1, 0.15) -- (1, -0.15); 

            \draw[dotted] (0,2) node[left] {$c$} -- (3,2);
            \draw[thick, blue!80!black] (-0.5,0) -- (3,0) -- (3,2) -- (4.2,2);
        \end{tikzpicture}
        \caption{$\operatorname{stepAtEnd}(x, c)$}
    \end{subfigure}

    \vspace{0.8cm}


    \begin{subfigure}{0.32\textwidth}
        \centering
        \begin{tikzpicture}[scale=0.75]
            \draw[->] (-0.5,0) -- (4.5,0) node[right] {$t$};
            \draw[->] (0,-1.5) -- (0,3) node[above] {};

            \filldraw[fill=blue!40!white, draw=black] (1,-1) rectangle (3,-0.2);
            \node at (2,-0.6) {$x$};

            \draw[fill=white, draw=black] (-0.2, 1) rectangle (0.2, 2.5);
            \filldraw[black] (0,1.75) circle (1.5pt);
            \node[left] at (-0.2, 1) {$\underline{c}$};
            \node[left] at (-0.2, 2.5) {$\overline{c}$};

            \draw[dotted] (0,1.75) -- (1,1.75);
            \draw[thick, blue!80!black] (-0.5,0) -- (1,0) -- (1,1.75) -- (3,1.75) -- (3,0) -- (4.2,0);
        \end{tikzpicture}
        \caption{$\operatorname{pulse}(x, \underline{c}, \overline{c})$}
    \end{subfigure}\hfill
    \begin{subfigure}{0.32\textwidth}
        \centering
        \begin{tikzpicture}[scale=0.75]
            \draw[->] (-0.5,0) -- (4.5,0) node[right] {$t$};
            \draw[->] (0,-1.5) -- (0,3) node[above] {};

            \filldraw[fill=blue!40!white, draw=black] (1,-1) rectangle (3,-0.2);
            \node at (2,-0.6) {$x$};
            \draw (3, 0.15) -- (3, -0.15); 

            \draw[fill=white, draw=black] (-0.2, 1) rectangle (0.2, 2.5);
            \filldraw[black] (0,1.75) circle (1.5pt);
            \node[left] at (-0.2, 1) {$\underline{c}$};
            \node[left] at (-0.2, 2.5) {$\overline{c}$};

            \draw[dotted] (0,1.75) -- (1,1.75);
            \draw[thick, blue!80!black] (-0.5,0) -- (1,0) -- (1,1.75) -- (4.2,1.75);
        \end{tikzpicture}
        \caption{$\operatorname{stepAtStart}(x, \underline{c}, \overline{c})$}
    \end{subfigure}\hfill
    \begin{subfigure}{0.32\textwidth}
        \centering
        \begin{tikzpicture}[scale=0.75]
            \draw[->] (-0.5,0) -- (4.5,0) node[right] {$t$};
            \draw[->] (0,-1.5) -- (0,3) node[above] {};

            \filldraw[fill=blue!40!white, draw=black] (1,-1) rectangle (3,-0.2);
            \node at (2,-0.6) {$x$};
            \draw (1, 0.15) -- (1, -0.15); 

            \draw[fill=white, draw=black] (-0.2, 1) rectangle (0.2, 2.5);
            \filldraw[black] (0,1.75) circle (1.5pt);
            \node[left] at (-0.2, 1) {$\underline{c}$};
            \node[left] at (-0.2, 2.5) {$\overline{c}$};

            \draw[dotted] (0,1.75) -- (3,1.75);
            \draw[thick, blue!80!black] (-0.5,0) -- (3,0) -- (3,1.75) -- (4.2,1.75);
        \end{tikzpicture}
        \caption{$\operatorname{stepAtEnd}(x, \underline{c}, \overline{c})$}
    \end{subfigure}

    \caption{Elementary cumulative function expressions with fixed consumption $c$ (top row) and variable consumption range $[\underline{c}, \overline{c}]$ (bottom row) over interval $x$.}
    \label{fig:elementary}
\end{figure}

\begin{definition}[Elementary Cumulative Functions]
\label{def:elementary}
The following elementary cumulative functions~\cite{LaborieRSV09} express the contribution of a conditional time-interval variable~$x$ (with execution status $p_x$, start $s_x$, end $e_x$) and a fixed consumption $c \in \mathbb{Z}^+$ to a step-wise resource profile:
\begin{align*}
  \mathit{pulse}(x,c)(t) &= \begin{cases} c & \text{if } p_x = \mathit{true} \text{ and } s_x \leq t < e_x \\ 0 & \text{otherwise} \end{cases}\\
  \mathit{stepAtStart}(x,c)(t) &= \begin{cases} c & \text{if } p_x = \mathit{true} \text{ and } t \geq s_x \\ 0 & \text{otherwise} \end{cases}\\
  \mathit{stepAtEnd}(x,c)(t) &= \begin{cases} c & \text{if } p_x = \mathit{true} \text{ and } t \geq e_x \\ 0 & \text{otherwise} \end{cases}
\end{align*}
All three functions are step-wise integer functions that do not vary within an interval.
When the consumption ranges over $[\underline{c}, \overline{c}]$ with $\underline{c} \leq \overline{c}$ and $\underline{c}, \overline{c} \in \mathbb{Z}^+$, the functions are defined analogously with a consumption variable $c \in [\underline{c}, \overline{c}]$.
Figure~\ref{fig:elementary} illustrates both variants (fixed consumption in the top row, consumption range in the bottom row) for each of the three functions.
A deterministic step function for a fixed time point $t_0 \in \mathbb{Z}$ and $c \in \mathbb{Z}$ is also supported:
\[
  \mathit{step}(t_0,c)(t) = \begin{cases} c & \text{if } t \geq t_0 \\ 0 & \text{otherwise.} \end{cases}
\]
\end{definition}

Cumulative functions (elementary or not) can be composed using two binary operators.
Given two cumulative functions~$f$ and $g$, the \emph{plus} and \emph{minus} operators are defined point-wise as:
\[
  \mathit{plus}(f,g)(t) = f(t) + g(t), \qquad \mathit{minus}(f,g)(t) = f(t) - g(t).
\]
A composite cumulative function can therefore be viewed as an Abstract Syntax Tree (AST) with elementary functions ($\mathit{pulse}$, $\mathit{stepAtStart}$, $\mathit{stepAtEnd}$, $\mathit{step}$) at the leaf nodes and $\mathit{plus}$/$\mathit{minus}$ nodes as internal operators.
Flattening a composite cumulative function produces a multiset of cumulative tasks $(x',c')$ where $c'\in \mathbb{Z}$ is an integer consumption and $x'$ is either the original interval (for $\mathit{pulse}$) or a fresh interval suitably constrained to the original interval (for $\mathit{stepAtStart}$ and $\mathit{stepAtEnd}$).
\begin{definition}[Cumulative Task]
\label{def:cumulative-task}
A \emph{cumulative task} is a pair $(x,c)$ where $x \in I$ is a conditional time-interval variable and $c \in \mathbb{Z}$ is its consumption on the resource. We denote by $T$ the set of cumulative tasks (or \emph{tasks} for short) in the problem.
\end{definition}

\begin{example}
\label{ex:cumfun}
Let~$A$, $B$, and $C$ be three mandatory (with execution status true) conditional time-interval variables with fixed start times and durations (depicted as double-headed arrows in Figure~\ref{fig:example}, top right).
They contribute to a cumulative function $f = \mathit{stepAtStart}(A,2) \allowbreak - \allowbreak(\mathit{pulse}(B,1) \allowbreak + \allowbreak \mathit{stepAtEnd}(C,1))$.
The resulting AST is shown on the left of Figure~\ref{fig:example}.
On the bottom right, the colored border rectangles are the cumulative tasks $\{(A',2),(B,-1),(C',-1)\}$ obtained by flattening~$f$ (see Algorithm~\ref{algo:flatten}); and the dotted step function is the resulting cumulative function~$f$.
\begin{figure}[tbp]
    \centering
    \includegraphics[width=0.85\linewidth]{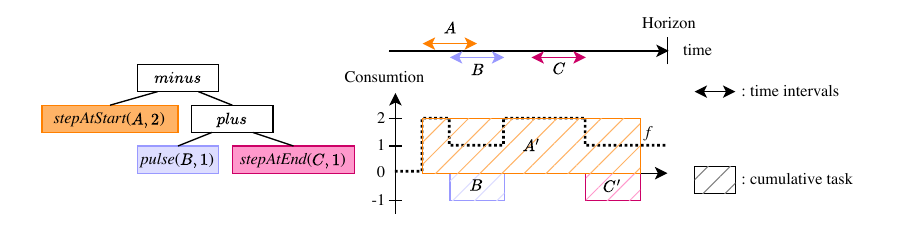}
    \caption{Example~\ref{ex:cumfun}: AST of $f$ (left) and resource profile (right). Double-headed arrows (top) show the fixed time-intervals of $A$, $B$, $C$. Colored border rectangles represent the cumulative tasks $\{(A',2),(B,-1),(C',-1)\}$ obtained by flattening~$f$; the dotted line is the resulting cumulative function~$f$.}
    \label{fig:example}
\end{figure}
\end{example}

A cumulative function can be flattened using Algorithm~\ref{algo:flatten} to extract a set of cumulative tasks, each represented as a pair $(x, c)$ of an interval~$x$ and a consumption~$c$ (positive or negative). This algorithm traverses the AST and collects the intervals at the leaf nodes, reversing the consumption signs for the intervals flattened on the right branch of a minus node.
Note that whenever encountering a $\mathit{stepAtStart}(x, c)$ leaf node, the cumulative task collected is defined on a fresh interval~$x'$ related to the original one with: $x'.s = x.s$ (same start as $x$), $x'.p = x.p$ (same execution status), and $x'.e = Horizon$ (ending at the time horizon) to reflect the definition of a step function.
Similarly, in the case of $\mathit{stepAtEnd}(x, c)$, a new interval~$x'$ with $x'.s = x.e$ (same start as end of $x$) is collected.

\begin{algorithm}[tbp]
    \DontPrintSemicolon
    \SetKwProg{Fn}{Function}{:}{}
    \KwIn{$n$: A node of the AST}
    \KwOut{Set of cumulative tasks}
    \lIf{$n = \mathit{pulse}(x, c)$}{
        \Return $\{(x, c)\}$
    }
    \ElseIf{$n = \mathit{stepAtStart}(x, c)$}{
        $x' \gets \langle x.s, Horizon - x.s, Horizon, x.p \rangle$ \tcp*{Fresh interval for stepAtStart}
        \Return $\{(x', c)\}$
    }
    \ElseIf{$n = \mathit{stepAtEnd}(x, c)$}{
        $x' \gets \langle x.e, Horizon - x.e, Horizon, x.p \rangle$ \tcp*{Fresh interval for stepAtEnd}
        \Return $\{(x', c)\}$
    }
    \lElseIf{$n = \mathit{plus}(n_l, n_r)$}{
        \Return $\mathit{flatten}(n_l) \cup \mathit{flatten}(n_r)$
    }
    \lElseIf{$n = \mathit{minus}(n_l, n_r)$}{
        \Return $\mathit{flatten}(n_l) \cup \{(x, -c) \mid (x, c) \in \mathit{flatten}(n_r)\}$
    }
    \ElseIf{$n = \mathit{step}(t_0, c)$}{
        $x' \gets \langle t_0, Horizon - t_0, Horizon, \mathit{true} \rangle$ \tcp*{Fresh interval for step}
        \Return $\{(x', c)\}$
    }
    \lElse{
        $error$ \tcp*[h]{Unknown pattern}
    }
    \caption{flatten($n$)}
    \label{algo:flatten}
\end{algorithm}

\begin{example}
\label{ex:profile-constraint}
Continuing example~\ref{ex:cumfun}, flatten($f$) collects the cumulative tasks $\{(A',2),(B,-1),(C',-1)\}$ illustrated on the right of Figure~\ref{fig:example}.
\end{example}

\subsection{Constraints on Cumulative Functions}

The height of a cumulative function can be restricted with a lower and upper bound at every time overlapping at least one executing time-interval: $\minC \leq f \leq \maxC$.
This constraint is called $alwaysIn(f,\minC,\maxC)$ in~\cite{LaborieRSV09}.
It can be enforced by posting a \gcumulative\ constraint requiring a resource consumption between $\minC$ and $\maxC$ for the collection of cumulative tasks $T = \mathit{flatten}(f)$ (treated as a multiset preserving distinct task instances with identical parameters).
The whole domain of a cumulative task~$i \in T$ is encoded by a tuple of variables~$\langle s_i, d_i, e_i, c_i, p_i \rangle$ where the integer variable~$c_i$ represents its consumption on the resource; $c_i$ may be positive (producer), negative (consumer), or its domain $[\minc_i, \maxc_i]$ may span both negative and positive values (i.e., $\minc_i < 0 < \maxc_i$).
The set of tasks~$T$ is partitioned into three subsets as given in Definition~\ref{def:subsets}.

\begin{definition}
\label{def:subsets}
The set $O = \{i\in T\mid p_i \in \{\mathit{true}, \mathit{false}\}\}$ denotes the set of optional tasks, those for which it is not decided yet if they will be executed or not on the resource.
The set $R=\{i\in T\mid p_i=\mathit{true}\}$ denotes the set of required tasks, those which will surely be executed on the resource.
The set $E=\{i\in T\mid p_i=\mathit{false}\}$ denotes the set of excluded tasks, those which will surely not be executed on the resource.
The sets~$O$, $R$ and $E$ are disjoint and form a partition of the set~$T$, i.e., $T=O\cup R\cup E$.
A task~$i$ is \emph{fixed} when its execution is required ($p_i = \mathit{true}$), its execution window is fixed ($\mins_i = \maxs_i, \mine_i = \maxe_i$), and its consumption is fixed ($\minc_i = \maxc_i$); otherwise, it is \emph{non-fixed}.
\end{definition}  


\begin{definition}
    The $\gcumulative(T,\minC,\maxC)$ constraint enforces the cumulated consumptions of the tasks of $T$ to be within the fixed capacity range $[\minC, \maxC]$ for every time at which at least one task of $T$ is executed:
\begin{gather}
    \sum_{i \in T \mid p_i = \mathit{true} \wedge s_i \leq \tau < e_i} c_i \in [\minC, \maxC] \qquad \forall \tau \in \bigcup_{i \in T \mid p_i = \mathit{true}} [s_i,e_i)
    \label{eq:decomp}
\end{gather}
\end{definition}

\paragraph{Restricted Time Interval Scope} To enforce the capacity range at every time point across the entire horizon rather than only when at least one task executes, one can add a dummy task that spans the entire horizon, with a mandatory execution status and a consumption of zero.
More generally, the time scope can also be limited to a specific task, by linking the status variables of the tasks to the overlap of this task.

\paragraph{Consistency} Enforcing global bound-consistency for the cumulative constraint is NP-Hard~\cite{GareyJ79}.
However, enforcing it on the decomposition into sum constraints at every point in the horizon, as in~\eqref{eq:decomp}, is not NP-Hard.
Unfortunately, this approach is inefficient for large horizons, as it requires a sum constraint to be enforced at every time point.
Therefore, we aim to achieve the same filtering in polynomial time relative to the number of tasks, rather than the size of the horizon.
This filtering is generally referred to as \textit{timetabling} filtering.
Intuitively, it works by aggregating all the fixed parts (when $\maxs_i < \mine_i$) to create a pessimistic consumption profile.
It then postpones the earliest execution of a task whenever it detects that starting it at this time would overload the resource capacity.

The next examples illustrate the modeling of two classical cumulative scheduling scenarios with cumulative functions and the \gcumulative\ constraint.

\begin{example}[Classical cumulative constraint]
\label{ex:classical-cumul}
Consider $n$ tasks $T = \{(x_i,c_i)\mid i = 1,\ldots,n\}$, each with a fixed positive consumption $c_i > 0$, a mandatory execution ($p_i = \mathit{true}$), and a resource of capacity $\mathcal{C} > 0$.
The classical \textit{cumulative} constraint requires that the total consumption of concurrently executing tasks never exceeds $\mathcal{C}$.
This is captured by the cumulative function
\[
    f = \sum_{i=1}^{n} \mathit{pulse}(x_i, c_i),
\]
and posting $\gcumulative(\mathit{flatten}(f), -\infty, \mathcal{C})$ enforces that the total consumption stays under $\mathcal{C}$ at all times at which at least one task executes.
Note that since $f$ contains only $\mathit{pulse}$ terms combined with $+$, the \textit{flatten} function simply collects the pairs $(x_i, c_i)$ without any sign reversal --- it is essentially the identity in this case.
\end{example}

\begin{example}[Producer/consumer constraint]
\label{ex:producer-consumer}
Consider a reservoir (e.g., a tank or a battery) with level required to stay within $[L_{\min}, L_{\max}]$ at all times.
A set of tasks $T = \{(x_i,c_i)\mid i = 1,\ldots,n\}$ is defined, each with a fixed consumption~$c_i$ (positive or negative).
Tasks that \emph{produce} resources add to the level (positive consumption), while tasks that \emph{consume} resources subtract from it (negative consumption).
Let~$P \subseteq T$ be the set of producer tasks, each with consumption $c_i > 0$, and $C \subseteq T$ be the set of consumer tasks, each with consumption $-c_j < 0$ ($c_j > 0$).
Modeling the initial level $\ell_0$ of the reservoir at time~$0$ can be done with a $\mathit{step}(0, \ell_0)$ function.
The cumulative function
\[
    f = \mathit{step}(0, \ell_0) + \sum_{i \in P} \mathit{stepAtStart}(x_i, c_i) - \sum_{j \in C} \mathit{stepAtStart}(x_j, c_j)
\]
represents the reservoir level over time.
Posting $\gcumulative(\mathit{flatten}(f), L_{\min}, L_{\max})$ enforces that the level stays within $[L_{\min}, L_{\max}]$ at all times at which at least one task executes.
\end{example}

\section{Timetabling Algorithm}\label{sec:timetabling}

This section first introduces the \textit{Profile} data structure then explains the timetabling filtering algorithm for filtering the attributes of cumulative tasks.

\subsection{The Profile Data Structure}
The \textit{Profile} data structure is an aggregation of adjoined rectangles of different durations and heights introduced by~\textcite{GingrasQ16} and~\textcite{GayHS15} to record the resource utilization of tasks.
The end of a rectangle is the beginning of the next one.
The starting (resp. ending) of a rectangle is represented by a tuple called \emph{time point} with a component representing the beginning (resp. ending) time of the rectangle and other components that contain information relative to the resource consumption for the duration of the rectangle.
These time points are sorted in increasing order of time and are kept in a doubly-linked list structure called the \textit{Profile}, bounded by sentinel head and tail nodes with profile values of zero.
An original idea of~\textcite{GingrasQ16} is to have pointers $\mathit{tpAt}(\mins_i)$ and $\mathit{tpAt}(\maxe_i)$ linking to the time point associated with $\mins_i$ and $\maxe_i$ respectively.

\subsection{Minimum and Maximum Profiles}

To check and adjust the attributes of tasks, the \textit{Profile} data structure from~\textcite{GingrasQ16} is extended to compute both a minimum and maximum profile range at each time point.
In our adaptation of the \textit{Profile} data structure (hereafter referred to as \textit{timeline} and noted $\mathcal{L}$), a time point $tp \in \mathcal{L}$ consists of a tuple $\langle \mathit{time}, \minP, \maxP, \mathit{\#fp} \rangle$, where $\mathit{time}$ corresponds to the start time of the time point, $\minP$ (resp. $\maxP$) to the minimum (resp. maximum) resource profile of tasks over time and $\mathit{\#fp}$ to the number of fixed parts of tasks that overlap over $\mathit{time}$.

Intuitively, the minimum profile is obtained by considering the upper bound of the negative contribution and the lower bound of the positive contribution of each task.
In contrast, the maximum profile considers the upper bound of the positive contribution and the lower bound of the negative contribution of each task.
The upper bound of the negative (resp. positive) contribution of a task corresponds to its largest negative consumption in absolute terms, i.e., minimal algebraic value $\min(\minc_i, 0)$ (resp. upper bound $\max(\maxc_i, 0)$), during the whole possible time window of the task,
while the lower bound of its negative (resp. positive) contribution corresponds to its smallest negative consumption in absolute terms, i.e., $\min(\maxc_i, 0)$ (resp. lower bound $\max(\minc_i, 0)$), during the fixed part of the task. More formally, the minimum (resp. maximum) resource profile at time point~$tp$ denoted  $tp.\minP$ (resp. $tp.\maxP$) is defined as:
\begin{gather}
    tp.\minP = \sum\limits_{i \in O \cup R \mid \mins_i \leq tp.\mathit{time} < \maxe_i} \min(\minc_i, 0) + \sum\limits_{i \in R \mid \maxs_i \leq tp.\mathit{time} < \mine_i} \max(\minc_i, 0), \label{eq:minP} \\
    tp.\maxP = \sum\limits_{i \in O \cup R \mid \mins_i \leq tp.\mathit{time} < \maxe_i} \max(\maxc_i, 0) + \sum\limits_{i \in R \mid \maxs_i \leq tp.\mathit{time} < \mine_i} \min(\maxc_i, 0), \label{eq:maxP}
\end{gather}
where $\{ i \in O \cup R \mid \mins_i \leq tp.\mathit{time} < \maxe_i \}$ corresponds to the set of all optional and required tasks whose time window overlaps the time $tp.\mathit{time}$ and $\{ i \in R \mid \maxs_i \leq tp.\mathit{time} < \mine_i \}$ corresponds to the set of all required tasks that have a fixed part that overlaps the time $tp.\mathit{time}$.

The proposed timetabling filtering algorithm is split into two steps:
First, the timeline~$\mathcal{L}$ is initialized by creating and ordering the time points, linking them with the tasks, computing their minimum and maximum profile ranges, and checking the feasibility consistency by comparing the profile ranges with the maximum and minimum capacity.
Second, each task is checked during its entire possible span in terms of profile and capacity to adjust its domain.

\subsection{Timeline Initialization and Feasibility Consistency Check}
The function \textit{initializeTimeline} (Algorithm~\ref{algo:init}) initializes the timeline, links the corresponding time points to tasks, and computes the minimum and maximum profile of the resource utilization.
It receives as input the set of tasks~$T$ and the bounds of the resource capacity $[\minC, \maxC]$.
First, events are generated by iterating over the tasks that are not absent (Line~\ref{algloop:events}).
An event associated with a task~$i$ is a triple $e=\langle\mathit{time}, \mathit{type}, \mathit{task}\rangle$ where $e.\mathit{time}\in \{\mins_i, \maxs_i, \mine_i, \maxe_i \mid i\in T\}$ is the time of the event, $e.\mathit{type}\in \{\mins, \maxs, \mine, \maxe\}$ is the type of the event (start min, start max, end min, or end max), and $e.task=i$ is the index of the task.
Each non-absent task generates a maximum of four events, two for the start and end of its possible time window and two for the start and end of its fixed part if it has one.
For each non-absent task~$i$, two events are generated which correspond to the minimum start time $(\mins_i, \mins, i)$, and the maximum end time $(\maxe_i, \maxe, i)$ of the task.
For tasks that are present and with a fixed part, two additional events are generated for the maximum start time $(\maxs_i, \maxs, i)$ and the minimum end time $(\mine_i, \mine, i)$ of the task.

A current time point~$tp$ is maintained and corresponds to the last time point of the timeline being initialized.
Then, these events are processed by order of increasing time to initialize the timeline~$\mathcal{L}$ (Line~\ref{initializeTimeline}).
New time points are created only when the time of the event differs from the current time point~$tp$ as checked at Line~\ref{algline:newTP}.
This means that if several events share the same time, they will all contribute towards the same time point; because accumulator additions and subtractions are commutative, any tie-breaking order among events that share the same time yields the identical time point profile.
When a new time point is created, it is appended at the end of the timeline which links it to the previous last time point.
When processing a start min ($\mins$) or end max ($\maxe$) event, the current time point is linked to the task event at Lines~\ref{algline:linkTP1} and~\ref{algline:linkTP2} to retrieve the time points corresponding to the start and end of tasks in constant time later.

Three accumulators are used to keep track of the minimum ($\minP$) and maximum ($\maxP$) profile as well as the number of fixed parts of tasks overlapping the current time point ($\mathit{\#fp}$).
Depending on the nature of the event, the positive or negative contribution of its associated task is either added to (in case of $\mins$ or $\maxs$) or subtracted (in case of $\maxe$ or $\mine$) from the accumulators and the current time point~$tp$ is updated.

As the timeline is initialized, the feasibility consistency of the minimum and maximum profiles is checked for each time point at Line~\ref{algline:constCheck}.
Once a time point is completed (when creating the next time point), we ensure that if at least one task is guaranteed to execute during the time point ($\mathit{\#fp} > 0$), the minimum profile $tp.\minP$ is not above the maximum capacity of the resource~$\maxC$ and the maximum profile $tp.\maxP$ is not below the minimum capacity of the resource~$\minC$.
At the last time point of the timeline, no task is guaranteed to execute, so the consistency check is not performed.
The correctness of the profile construction and the soundness of the infeasibility check are established in the following lemma and theorem (proofs are deferred to Appendix~\ref{sec:proofs}).
\begin{lemma}[Correctness of the profile bounds]
\label{lem:profile-bounds}
For every time point $tp \in \mathcal{L}$, immediately after processing all events whose time is at most $tp.\mathit{time}$, the accumulator values $tp.\minP$ and $tp.\maxP$ satisfy~\eqref{eq:minP} and~\eqref{eq:maxP}, respectively, and the number of overlapping fixed parts is exactly
\[
    tp.\#fp = \left|\{ i \in R \mid \maxs_i \leq tp.\mathit{time} < \mine_i \}\right|.
\]
\end{lemma}

\begin{theorem}[Soundness of the infeasibility check]
\label{thm:infeasibility}
If Algorithm~\ref{algo:init} returns \texttt{Fail} at Line~\ref{algline:fail}, then the current domains of the tasks are inconsistent with the \gcumulative \ constraint.
\end{theorem}

The complexity of the \textit{initializeTimeline} function is given in Theorem~\ref{thm:init-complexity}.
\begin{theorem}[Complexity of \texttt{initializeTimeline}]
\label{thm:init-complexity}
Algorithm~\ref{algo:init} has a worst-case time complexity of~$\mathcal{O}(n \log n)$, where $n = |T|$.
\end{theorem}

Example~\ref{ex:profile} illustrates the profile initialization for three tasks.

\begin{algorithm}[tbp]
    \DontPrintSemicolon
    $\mathit{events} \gets \{\}$ \tcp*{Initialize the set of events}
    \For{$i \in \{j \in T \mid p_j \neq \mathit{false}\}$}{\label{algloop:events}
        $\mathit{events} \gets \mathit{events} \cup \{(\mins_i, \mins, i),(\maxe_i, \maxe, i)\}$ \tcp*{Each event is a (time, type, task) triple}
        \If{$p_i = \mathit{true} \wedge \maxs_i < \mine_i$}{
            $\mathit{events} \gets \mathit{events} \cup \{(\maxs_i, \maxs, i),(\mine_i, \mine, i)\}$ \tcp*{Add fixed-part events for required tasks}
        }
    }
    $\minP \gets 0$; $\maxP \gets 0$  \tcp*{Lower/Upper bound of the profile at current time point}
    $\mathit{\#fp} \gets 0$ \tcp*{Number of overlapping fixed parts of current time point}
    $time \gets \min_{e \in \mathit{events}} e.\mathit{time}$ \tcp*{Initialize current time to the earliest event time}
    $tp \gets \langle time,\minP, \maxP, \mathit{\#fp} \rangle$ \tcp*{Current time point; a timepoint record stores $\{\mathit{time}, \minP, \maxP, \mathit{\#fp}\}$}
    $\mathcal{L} \gets \{tp\}$\label{initializeTimeline} \tcp*{All time points created so-far}
    
    \For{$e \in \mathit{events}$ sorted by time} {\label{algloop:setDeltas}
        $i \gets e.task$\;
        \If{$tp.\mathit{time} < e.\mathit{time}$}{ \label{algline:newTP}
            \If{$\mathit{\#fp} > 0 \wedge (\minP > \maxC \vee \maxP < \minC)$}{\label{algline:constCheck}
                \Return Fail \tcp*{Capacity violation detected}\label{algline:fail}
            }
            $tp \gets \langle e.\mathit{time},\minP, \maxP, \mathit{\#fp} \rangle$ \tcp*{New time point}
            $\mathcal{L}.append(tp)$\;
        }
        
        \If(\tcp*[f]{Process earliest start event}){$e.\mathit{type} = \mins$}{ \label{algline:deltaInc}
            $\minP \gets \minP + \min(\minc_i,0)$ \tcp*{Add pessimistic negative contribution}
            $\maxP \gets \maxP + \max(\maxc_i,0)$\ \tcp*{Add optimistic positive contribution}
            $\mathit{tpAt}(\mins_i) \gets tp$ \tcp*{Store time-point pointer for $\mins_i$}\label{algline:linkTP1}
        }

        \If(\tcp*[f]{Process latest end event}){$e.\mathit{type} = \maxe$}{ \label{algline:deltaDec}
            $\minP \gets \minP - \min(\minc_i,0)$ \tcp*{Remove pessimistic negative contribution}
            $\maxP \gets \maxP - \max(\maxc_i,0)$\ \tcp*{Remove optimistic positive contribution}
            $\mathit{tpAt}(\maxe_i) \gets tp$ \tcp*{Store time-point pointer for $\maxe_i$}\label{algline:linkTP2}
        }
        
        \If{$p_i = \mathit{true} \wedge \maxs_i < \mine_i$}{
            \If(\tcp*[f]{Process start of fixed parts event}){$e.\mathit{type} = \maxs$}{ \label{algline:deltaFp1}
                $\minP \gets \minP + \max(\minc_i,0)$ \tcp*{Add pessimistic positive contribution}
                $\maxP \gets \maxP + \min(\maxc_i,0)$\ \tcp*{Add optimistic negative contribution}
                $\mathit{\#fp} \gets \mathit{\#fp} + 1$ \tcp*{Increment overlapping fixed part counter}
            }
            \If(\tcp*[f]{Process end of fixed parts event}){$e.\mathit{type} = \mine$}{\label{algline:deltaFp2}
                $\minP \gets \minP - \max(\minc_i,0)$  \tcp*{Remove pessimistic positive contribution}
                $\maxP \gets \maxP - \min(\maxc_i,0)$ \tcp*{Remove optimistic negative contribution}
                $\mathit{\#fp} \gets \mathit{\#fp} - 1$ \tcp*{Decrement overlapping fixed part counter}
            }
        }
        $tp.\minP \gets \minP$; $tp.\maxP \gets \maxP$; $tp.\mathit{\#fp} \gets \mathit{\#fp}$ \tcp*{Update current time point attributes}
    }
    \Return $\mathcal{L}$ \tcp*{Return the complete timeline}
    \caption{initializeTimeline($T,\minC, \maxC$)}
    \label{algo:init}
\end{algorithm}

\begin{example}
    \label{ex:profile}
    Let us consider three tasks :
    \begin{description}
        \item[$A$] : $\langle
            s = [0,1],\allowbreak
            d = [3,4],\allowbreak
            e = [3,4],\allowbreak
            c = [1,2],\allowbreak
            p = \mathit{true}
        \rangle$;
        \item[$B$] : $\langle
            s = [2,4],\allowbreak
            d = [3,4],\allowbreak
            e = [5,7],\allowbreak
            c = 2,\allowbreak
            p = \mathit{true}
        \rangle$;
        \item[$C$] : $\langle
            s = [3,8],\allowbreak
            d = [1,3],\allowbreak
            e = [4,9],\allowbreak
            c = [-2,1],\allowbreak
            p = \{\mathit{true},\mathit{false}\}
        \rangle$;
    \end{description}
    and a resource of capacity bounds $[0,1]$.
    The time points and their attributes are shown in Table~\ref{tab:profile}.
    \begin{table}[tbp]
        \centering
        \begin{tabular}{|c|l|ccc|}
        \hline
        time & \multicolumn{1}{c|}{event(s)} & $\minP$ & $\maxP$ & $\mathit{\#fp}$ \\ \hline
        0    & $\mins_A$            & 0    & 2    & 0        \\
        1    & $\maxs_A$            & 1    & 2    & 1        \\
        2    & $\mins_B$            & 1    & 4    & 1        \\
        3    & $\mine_A, \mins_C$   & -2   & 5    & 0        \\
        4    & $\maxe_A, \maxs_B$   & 0    & 3    & 1        \\
        5    & $\mine_B$            & -2   & 3    & 0        \\
        7    & $\maxe_B$            & -2   & 1    & 0        \\
        9    & $\maxe_C$            & 0    & 0    & 0        \\ \hline
        \end{tabular}
        \caption{Time points computed from the tasks of Example~\ref{ex:profile}.}
        \label{tab:profile}
    \end{table}
    
    Note that the events~$\maxs_C$ and $\mine_C$ are not considered because task~$C$ is optional and has no fixed part ($\maxs_C \geq \mine_C$).
    Figure~\ref{fig:profile} shows a representation of the tasks and the resulting profile ranges.
    \begin{figure}[tbp]
        \centering
        \includegraphics[width=0.68\linewidth]{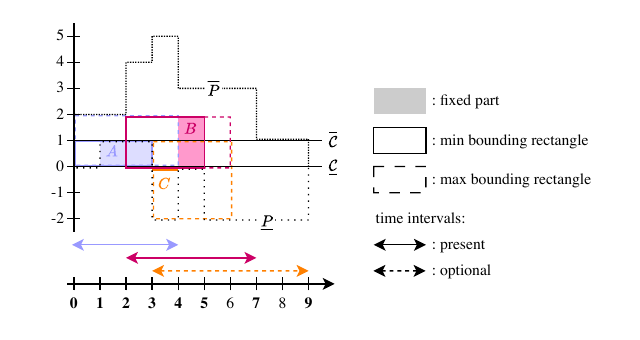}
        \caption{Representation of the tasks of Example~\ref{ex:profile} with the resulting minimum ($\minP$) and maximum ($\maxP$) profile. The minimum and maximum bounding rectangles of each task are also represented.}
        \label{fig:profile}
    \end{figure}
\end{example}

\subsection{Filtering}
\label{subsec:filtering}

We first introduce a formal notion of timetabling filtering consistency for the \gcumulative \ constraint, and then present the four filtering rules whose purpose is to achieve it.
Recall that a task~$i \in T$ is fixed when $p_i = \mathit{true}$, $\mins_i = \maxs_i, \mine_i = \maxe_i$, and $\minc_i = \maxc_i$ (Definition~\ref{def:subsets}); otherwise, it is non-fixed.
Timetabling filtering operates with respect to the timeline profile $\mathcal{L}$ constructed once per propagation cycle by Algorithm~\ref{algo:init}.
When $\mathcal{L}$ is computed from the initial task domains, tasks are partitioned according to their presence status at that moment into initially required tasks $R_{\text{init}} = \{i \in T \mid p_i = \mathit{true}\}$ and initially optional tasks $O_{\text{init}} = \{i \in T \mid p_i \in \{\mathit{true}, \mathit{false}\}\}$.
Crucially, in Algorithm~\ref{algo:init} (Line~\ref{algline:deltaFp1}), only tasks that are already required at initialization ($i \in R_{\text{init}}$) have their fixed part compiled into the profile $\mathcal{L}$; optional tasks do not.
Because the timeline~$\mathcal{L}$ remains static throughout the filtering pass, whether a task's fixed part is embedded in $\mathcal{L}$ is determined once and for all by its membership in $R_{\text{init}}$ or $O_{\text{init}}$, even if an optional task becomes mandatory during propagation.
This distinction is necessary to evaluate the residual capacity available to a task without double counting.

\begin{definition}[Conflicting Time Points]
\label{def:conflicting-tp}
Let us assume that the current task domains are feasibly consistent with the capacity constraints, let $\mathcal{L}$ be a timeline computed from the current task domains, and let $i \in R_{\text{init}} \cup O_{\text{init}}$ be a non-fixed task.
We denote by $\mathit{fp}_i$ the fixed part of task~$i$:
\[
    \mathit{fp}_i = \begin{cases}
        [\maxs_i, \mine_i) & \text{if } i \in R_{\text{init}} \text{ and } \maxs_i < \mine_i, \\
        \emptyset & \text{otherwise.}
    \end{cases}
\]
A time point $tp \in \mathcal{L}$ is said to be \emph{conflicting} with $i$ if it satisfies at least one of the following conditions:
\begin{enumerate}
    \item[\textbf{(C1)}] \emph{Forward-conflicting:} $tp.\mathit{time} \in [\mins_i, \min(\maxs_i,\mine_i))$ and
          \[
              tp.\minP + \max(\minc_i,0) > \maxC \quad \lor \quad tp.\maxP + \min(\maxc_i,0) < \minC.
          \]
    \item[\textbf{(C2)}] \emph{Backward-conflicting:} $tp.\mathit{time} \in [\max(\maxs_i,\mine_i), \maxe_i)$ and
          \[
              tp.\minP + \max(\minc_i,0) > \maxC \quad \lor \quad tp.\maxP + \min(\maxc_i,0) < \minC.
          \]
    \item[\textbf{(C3)}] \emph{Mandatory-conflicting:} $tp.\mathit{time} \in [\mins_i,\maxe_i) \setminus \mathit{fp}_i$ with $tp.\mathit{\#fp} > 0$, and
          \[
              tp.\minP - \min(\minc_i,0) > \maxC \quad \lor \quad tp.\maxP - \max(\maxc_i,0) < \minC.
          \]
    \item[\textbf{(C4)}] \emph{Overlap consumption-conflicting:} $\maxs_i < \mine_i$, $tp.\mathit{time} \in [\maxs_i, \mine_i)$, and
          \[
              \begin{cases}
                  \maxc_i > \maxC - (tp.\minP - \minc_i) \;\lor\; \minc_i < \minC - (tp.\maxP - \maxc_i), & \text{if } i \in R_{\text{init}} \text{ ($tp.\mathit{time} \in \mathit{fp}_i$)}, \\
                  \maxc_i > \maxC - (tp.\minP - \min(\minc_i,0)) \;\lor\; \minc_i < \minC - (tp.\maxP - \max(\maxc_i,0)), & \text{if } i \in O_{\text{init}}.
              \end{cases}
          \]
\end{enumerate}
Intuitively, conditions \textbf{(C1)} and \textbf{(C2)} identify time points where scheduling task~$i$ at its earliest start or latest end would violate capacity; \textbf{(C3)} identifies time points outside the fixed part where omitting task~$i$ causes an unavoidable capacity violation, requiring $i$ to be present; and \textbf{(C4)} restricts the consumption of task~$i$ over the interval $[\maxs_i, \mine_i)$ during which it necessarily executes if present. When $i \in R_{\text{init}}$, this interval is the fixed part $\mathit{fp}_i$ of task~$i$ and was compiled into the profile $\mathcal{L}$, so its full contribution is removed; when $i \in O_{\text{init}}$, task~$i$ has no fixed part in $\mathcal{L}$ ($\mathit{fp}_i = \emptyset$) and did not contribute to $\mathcal{L}$ with a fixed part, so only its non-fixed contribution is removed to avoid double counting.
\end{definition}

\begin{definition}[Timetabling Filtering Consistency]
\label{def:tt-consistency}
The constraint $\gcumulative(T,\minC,\maxC)$ is \emph{timetabling filtering consistent} with respect to $\mathcal{L}$ if every non-fixed task $i \in R_{\text{init}} \cup O_{\text{init}}$ satisfies the following conditions:
\begin{enumerate}
    \item[\textbf{(TC1)}] \emph{Pointwise consistency:} No time point $tp \in \mathcal{L}$ is conflicting with $i$ (of type (C1) through (C4) in Definition~\ref{def:conflicting-tp}).
    \item[\textbf{(TC2)}] \emph{Free-part consumption consistency:} If $\maxs_i \geq \mine_i$ and $\mind_i \geq 1$, its consumption bounds satisfy:
          \begin{align*}
              \maxc_i &\leq \max_{tp \in \mathcal{L} \mid \mine_i - 1 < tp.\mathit{next}.\mathit{time} \wedge tp.\mathit{time} \leq \maxs_i} \bigl(\maxC - (tp.\minP - \min(\minc_i,0))\bigr), \\
              \minc_i &\geq \min_{tp \in \mathcal{L} \mid \mine_i - 1 < tp.\mathit{next}.\mathit{time} \wedge tp.\mathit{time} \leq \maxs_i} \bigl(\minC - (tp.\maxP - \max(\maxc_i,0))\bigr).
          \end{align*}
    \item[\textbf{(TC3)}] \emph{Duration consistency:} If $\maxs_i \geq \mine_i$, its maximum duration satisfies $\maxd_i \leq \max_{[a,b) \in A} (b - a)$, where
          \[
              A = \left\{ [a,b) \subseteq [\mins_i,\maxe_i) \;\middle|\; \forall tp \in \mathcal{L} \text{ with } tp.\mathit{time} \in [a,b), \; tp \text{ is not (C1)- or (C2)-conflicting with } i \right\}.
          \]
\end{enumerate}
Intuitively, condition~\textbf{(TC1)} requires all local time-window, presence, and fixed-part consumption conflicts (Definition~\ref{def:conflicting-tp}) to be eliminated.
Condition~\textbf{(TC2)} applies to tasks without a fixed part ($\maxs_i \geq \mine_i$): because any execution of duration $\mind_i \geq 1$ necessarily intersects the core interval $[\mine_i - 1, \maxs_i]$, its consumption must be compatible with at least one point in this interval, bounding $[\minc_i, \maxc_i]$ by the available residual capacity.
Finally, condition~\textbf{(TC3)} reflects that task~$i$ cannot overlap any (C1)- or (C2)-conflicting time point; thus, every feasible execution must lie entirely within a conflict-free window $[a,b) \in A$, bounding the maximum duration~$\maxd_i$ by the longest such window.
\end{definition}

The four filtering rules below show how to enforce the conditions of Definitions~\ref{def:conflicting-tp} and~\ref{def:tt-consistency}:
the \textit{Forbid} rule (enforcing \textbf{(C1)} and \textbf{(C2)}), the \textit{Mandatory} rule (enforcing \textbf{(C3)}), and the overlap case of the \textit{Consumption} rule (enforcing \textbf{(C4)}) together achieve pointwise consistency \textbf{(TC1)};
the free-part case of the \textit{Consumption} rule achieves free-part consumption consistency \textbf{(TC2)};
and the \textit{Duration} rule achieves duration consistency \textbf{(TC3)}.

\paragraph{Forbid}
The first rule is used to reduce the time windows of tasks by checking for each task~$i \in T$ if starting (resp. ending) the task at its minimum start time (resp. maximum end time) is possible in regards to the profile range and the capacity range.
If this is not the case, the task is pushed to the earliest (resp. latest) time at which it can be placed without violating the capacity range.
Formally, the rule is defined as:
\begin{multline}
tp.\minP + \max(\minc_i, 0) > \maxC \;\vee\; tp.\maxP + \min(\maxc_i, 0) < \minC \implies \mins_i \geq tp.\mathit{next}.\mathit{time}, \\
\forall tp \in \mathcal{L} \text{ such that } \mins_i \leq tp.\mathit{time} < \min(\maxs_i, \mine_i)
\label{fb1}
\end{multline}
\begin{multline}
tp.\minP + \max(\minc_i, 0) > \maxC \;\vee\; tp.\maxP + \min(\maxc_i, 0) < \minC \implies \maxe_i \leq tp.\mathit{time}, \\
\forall tp \in \mathcal{L} \text{ such that } \max(\maxs_i, \mine_i) \leq tp.\mathit{time} < \maxe_i
\label{fb2}
\end{multline}

Note that as tasks are modeled with interval variables, the rule $e_i = s_i + d_i$ is enforced internally in the variable.
Thus, when a task~$i$ has its minimum start time~$\mins_i$ (resp. maximum end time~$\maxe_i$) adjusted, its minimum end time~$\mine_i$ (resp. maximum start time~$\maxs_i$) is also updated by the relation $\mine_i \geq \mins_i + \mind_i$ (resp. $\maxs_i \leq \maxe_i - \mind_i$) embedded in the interval variable.

\paragraph{Mandatory}
The second rule detects if the execution of a task~$i \in T$ at a time point~$tp$ is necessary to avoid a violation of the resource capacity range at this time point.
If this is the case, the task is adjusted to make sure it is present during the time point:
\begin{enumerate}
    \item if the task is optional, it is set to present;
    \item its time window is adjusted such that it executes over the whole time point;
    \item its consumption is adjusted to avoid the profile range violation.
\end{enumerate}
In formal terms, the rule is defined as:
\begin{equation}
\label{mand}
\begin{aligned}
    & tp.\minP - \min(\minc_i, 0) > \maxC \;\vee\; tp.\maxP - \max(\maxc_i, 0) < \minC \\
    & \quad \implies \begin{cases}
        p_i \gets \mathit{true}, \\
        \maxs_i \leq tp.\mathit{time}, \quad \mine_i \geq tp.\mathit{next}.\mathit{time}, \\
        \minc_i \geq \mathit{deficit} \quad (\text{if } \mathit{deficit} > 0), \\
        \maxc_i \leq \mathit{overload} \quad (\text{if } \mathit{overload} < 0)
    \end{cases} \\
    & \qquad \forall tp \in \mathcal{L} \text{ such that } tp.\mathit{time} \in [\mins_i, \maxe_i) \setminus \mathit{fp}_i \;\wedge\; tp.\mathit{\#fp} > 0
\end{aligned}
\end{equation}
where $\mathit{deficit} = \minC - (tp.\maxP - \max(\maxc_i, 0))$ and $\mathit{overload} = \maxC - (tp.\minP - \min(\minc_i, 0))$ are the deficit and overload of consumption needed for the profile range to intersect the capacity range.

\paragraph{Consumption}
This rule adjusts the consumption of each task~$i \in T$ in regard to the resource capacity range.
The maximum (resp. minimum) value of the consumption of a task is bounded by the difference between the maximum (resp. minimum) capacity of the resource and the minimum (resp. maximum) profile without the contribution of the task.
Two cases are possible:
\begin{enumerate}
    \item If the task has $\maxs_i < \mine_i$, its consumption is checked and adjusted directly at each time point in $[\maxs_i, \mine_i)$:
    \begin{align}
        & \begin{aligned}
            \maxc_i &\leq \maxC - (tp.\minP - \minc_i) \\
            \minc_i &\geq \minC - (tp.\maxP - \maxc_i)
        \end{aligned}
        && \forall tp \in \mathcal{L} : tp.\mathit{time} \in \mathit{fp}_i \quad (i \in R_{\text{init}}) \label{hei} \\[6pt]
        & \begin{aligned}
            \maxc_i &\leq \maxC - (tp.\minP - \min(\minc_i, 0)) \\
            \minc_i &\geq \minC - (tp.\maxP - \max(\maxc_i, 0))
        \end{aligned}
        && \forall tp \in \mathcal{L} : tp.\mathit{time} \in [\maxs_i, \mine_i) \quad (i \in O_{\text{init}}) \label{hei2}
    \end{align}
    If the task was initially required ($i \in R_{\text{init}}$), this interval is its fixed part $\mathit{fp}_i$ and contributed to the profile, so its full contribution is removed before the residual capacity is checked; if it was initially optional ($i \in O_{\text{init}}$), it has no fixed part in $\mathcal{L}$ ($\mathit{fp}_i = \emptyset$) and did not contribute to $\mathcal{L}$ with a fixed part, so only its non-fixed contribution is removed to avoid double counting.

    \item If the task has $\maxs_i \geq \mine_i$ (free part) and a strictly positive minimal duration ($\mind_i \geq 1$), the maximum available consumption over the minimum overlapping interval of the task is used to adjust its consumption:
    \begin{equation}
         \maxc_i \leq \max_{tp \in \{tp \in \mathcal{L} \mid \mine_i-1 < tp.\mathit{next}.\mathit{time} \wedge tp.\mathit{time} \leq \maxs_i\}} \bigl(\maxC - (tp.\minP - \min(\minc_i, 0))\bigr)
    \end{equation}
    \begin{equation}
        \minc_i \geq \min_{tp \in \{tp \in \mathcal{L} \mid \mine_i-1 < tp.\mathit{next}.\mathit{time} \wedge tp.\mathit{time} \leq \maxs_i\}} \bigl(\minC - (tp.\maxP - \max(\maxc_i, 0))\bigr)
    \end{equation}
    The minimum overlapping interval was introduced by~\textcite{gay2015time} and is defined for a task~$i \in T$ with $\mind_i \geq 1$ as the interval $[\mine_i - 1, \maxs_i]$. Intuitively, it corresponds to the smallest interval such that, no matter its start time, duration or end time, the task executes during at least one time unit of this interval. (If $\mind_i = 0$, the task may execute with duration zero without consuming any resource, so its consumption is not restricted by this rule.)
\end{enumerate}

\paragraph{Duration}
The last rule is inspired by the approach of~\textcite{ouelletProcessing}.
It adjusts the maximum duration~$\maxd$ of the tasks that have $\maxs_i \geq \mine_i$ (for tasks with $\maxs_i < \mine_i$, the adjustment would be redundant with the \textit{Forbid} rule).
The principle is to find the longest time span during which the task can be scheduled without a resource capacity violation.
Again, we rely on the internal propagation rules of the interval variables to set the task as absent if the adjustment would empty its domain and to update the $\mins_i$ and $\maxe_i$ attributes following the relations $\mins_i \geq \mine_i - \maxd_i$ and $\maxe_i \leq \maxs_i + \maxd_i$.
Formally, the rule is written as:

\begin{equation}
    \maxd_i \leq \max_{[a,b) \in A} (b - a), \label{eq:duration}
\end{equation}
where
\begin{equation*}
    A = \left\{ [a,b) \subseteq [\mins_i, \maxe_i) \;\middle|\;
    \begin{aligned}
        &\forall tp \in \mathcal{L} \text{ with } tp.\mathit{time} \in [a, b), \\
        &tp.\minP + \max(\minc_i,0) \leq \maxC \;\wedge\; tp.\maxP + \min(\maxc_i,0) \geq \minC
    \end{aligned}
    \right\}.
\end{equation*}
Intuitively, $A$ is the set of intervals where the task can be scheduled without underloading the minimum and overloading maximum capacity.

\paragraph{Algorithm}
The function \textit{timetabling} of Algorithm~\ref{algo:timetabling} enforces these rules for each task~$i \in T$.
This algorithm receives as input the set of tasks~$T$ and the bounds of the resource capacity $[\minC,\maxC]$.
The algorithm initializes the timeline with Algorithm~\ref{algo:init} and iterates over all non-fixed tasks at Line~\ref{algloop:timetabling}.
Each task is processed in three steps:

First (step 1), the \textit{Forbid} rule is used to adjust the minimum start time of the task by iterating forward over the time points in the interval $[\mins_i,\min(\maxs_i,\mine_i))$ in the loop at Line~\ref{algloop:ttfw}.
Second (step 2), the \textit{Forbid} rule is used to adjust the maximum end time of the task by iterating backward over the interval $[\max(\maxs_i, \mine_i), \maxe_i)$ in the loop at Line~\ref{algloop:ttbw}.

\begin{figure}[tbp]
    \centering
    \includegraphics[width=.5\linewidth]{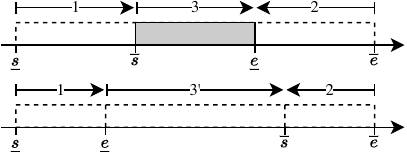}
    \caption{Timetabling processing order for a task with a fixed part (top) and one without (bottom). Those steps (1, 2, 3, and 3') are also referenced in Algorithm~\ref{algo:timetabling} comments.}
    \label{fig:timetabling}
\end{figure}

The last step iterates over the time points of the remaining part of the time window, depending on whether $\maxs_i < \mine_i$.
In the former case (step 3), the loop at Line~\ref{algloop:ttfp} iterates over $[\maxs_i,\mine_i)$: for tasks initially required ($i \in R_{\text{init}}$), this interval is its fixed part $\mathit{fp}_i$ and its consumption is adjusted following the 1st case of the \textit{Consumption} rule; for tasks initially optional ($i \in O_{\text{init}}$), it has no fixed part in $\mathcal{L}$, so \textit{checkIfMandatory} is evaluated at each time point to test whether the task must become mandatory.
In the latter case where $\maxs_i \geq \mine_i$ (step 3'), the loop at Line~\ref{algloop:ttfree} is executed.
The consumption of the task is adjusted according to the 2nd case of the \textit{Consumption} rule and its maximum duration is adjusted following the \textit{Duration} rule.
These three steps are illustrated in Figure~\ref{fig:timetabling}.

\begin{algorithm}[tbp]
    \DontPrintSemicolon
    $\mathcal{L} \gets \text{initializeTimeline}(T, \minC, \maxC)$ \tcp*{Build timeline profile from initial domains}
    $R_{\text{init}} \gets \{i\in T\mid p_i=\mathit{true}\}$;\quad $O_{\text{init}} \gets \{i\in T\mid p_i \in \{\mathit{true}, \mathit{false}\}\}$ \tcp*{Initial Required and Optional tasks}
    \For{$i \in R_{\text{init}} \cup O_{\text{init}}$}{ \label{algloop:timetabling}
        \lIf{$i$ is fixed $\vee \; \maxd_i = 0$} {continue \tcp*[f]{Skip fixed or 0-duration tasks} \label{algline:fixed}}
        $\mathit{tpf} \gets \mathit{tpAt}(\mins_i)$ \tcp*{First time point at or after $\mins_i$}
        \While(\tcp*[f]{Step~1: forward scan to adjust $\mins_i$}){$\mathit{tpf}.\mathit{time} < \min(\maxs_i, \mine_i)$}{ \label{algloop:ttfw}
            \lIf{$\mathit{tpf}.\minP + \max(\minc_i, 0) > \maxC \vee \allowbreak \mathit{tpf}.\maxP + \min(\maxc_i, 0) < \minC$}{
                $\mins_i \gets \max(\mathit{tpf}.\mathit{next}.\mathit{time}, \mins_i)$ \label{algline:fwadjust}
            }\lElse {
                $\text{checkIfMandatory}(i,\mathit{tpf}, \minC, \maxC)$ \tcp*[f]{Check mandatory condition (C3)} \label{algline:fwmand}
            }
            $\mathit{tpf} \gets \mathit{tpf}.\mathit{next}$\;
        }
        $\mathit{tpb} \gets \mathit{tpAt}(\maxe_i).\mathit{prev}$ \tcp*{Last time point before $\maxe_i$}
        \While(\tcp*[f]{Step~2: backward scan to adjust $\maxe_i$}){$\mathit{tpb}.\mathit{time} \geq \max(\maxs_i, \mine_i)$}{ \label{algloop:ttbw}
            \lIf{$\mathit{tpb}.\minP + \max(\minc_i, 0) > \maxC \vee \allowbreak \mathit{tpb}.\maxP + \min(\maxc_i, 0) < \minC$}{
                $\maxe_i \gets \min(\mathit{tpb}.\mathit{time}, \maxe_i)$ \label{algline:bwadjust}
            }\lElse{
                $\text{checkIfMandatory}(i,\mathit{tpb}, \minC, \maxC)$ \tcp*[f]{Check mandatory condition (C3)} \label{algline:bwmand}
            }
            $\mathit{tpb} \gets \mathit{tpb}.\mathit{prev}$\;
        }
        \If(\tcp*[f]{Step~3: task with fixed part ($\maxs_i < \mine_i$)}){$\maxs_i < \mine_i$}{
            $c_i^{\max} \gets \maxc_i$;\quad $c_i^{\min} \gets \minc_i$ \tcp*{Save bounds for this scan}
            \While(\tcp*[f]{Scan overlapping interval $[\maxs_i,\mine_i)$}){$\mathit{tpf}.\mathit{time} < \mine_i$}{ \label{algloop:ttfp}
                \lIf{$i \in O_{\text{init}}$}{$\text{checkIfMandatory}(i,\mathit{tpf}, \minC, \maxC)$ \label{algline:fpmand}}
                \If(\tcp*[f]{Required: fixed part contributed to $\mathcal{L}$}){$i \in R_{\text{init}}$}{
                    $\minc_i \gets \max(\minC - (\mathit{tpf}.\maxP - c_i^{\max}), \minc_i)$;\quad
                    $\maxc_i \gets \min(\maxC - (\mathit{tpf}.\minP - c_i^{\min}), \maxc_i)$ \label{algline:fph1}\;
                } \Else (\tcp*[f]{Optional: no fixed part in $\mathcal{L}$}) {
                    $\minc_i \gets \max(\minC - (\mathit{tpf}.\maxP - \max(c_i^{\max}, 0)), \minc_i)$;\quad
                    $\maxc_i \gets \min(\maxC - (\mathit{tpf}.\minP - \min(c_i^{\min}, 0)), \maxc_i)$ \label{algline:fph2}\;
                }
                $\mathit{tpf} \gets \mathit{tpf}.\mathit{next}$\;
            }
        } \Else (\tcp*[f]{Step~3': task without fixed part ($\maxs_i \geq \mine_i$), adjust $c_i, d_i$}) {
            $\maxd^* \gets 0$;\quad $s' \gets \mins_i$ \tcp*{Track feasible start and duration}
            \If(\tcp*[f]{Initialize bounds over core $[\mine_i-1, \maxs_i]$}){$\mind_i \geq 1$}{
                $\minc^* \gets \minC - \mathit{tpf}.\mathit{prev}.\maxP + \max(\maxc_i,0)$;\quad $\maxc^* \gets \maxC - \mathit{tpf}.\mathit{prev}.\minP + \min(\minc_i,0)$\;
            }
            \While(\tcp*[f]{Scan core interval $[\mine_i, \maxs_i)$}){$\mathit{tpf}.\mathit{time} < \maxs_i$}{\label{algloop:ttfree}
                $\maxd^* \gets \max(\mathit{tpf}.\mathit{time} - s', \maxd^*)$ \tcp*{Record conflict-free span}
                \lIf(\tcp*[f]{Reset start $s'$}){$\mathit{tpf}.\minP + \max(\minc_i, 0) > \maxC \vee \allowbreak \mathit{tpf}.\maxP + \min(\maxc_i, 0) < \minC$}{
                    $s' \gets \mathit{tpf}.\mathit{next}.\mathit{time}$ \label{algline:durationUpdate}
                }\lElse{
                    $\text{checkIfMandatory}(i,\mathit{tpf}, \minC, \maxC)$ \tcp*[f]{Check mandatory condition (C3)} \label{algline:freemand}
                }
                \If{$\mind_i \geq 1$}{
                    $\minc^* \gets \min(\minC - \mathit{tpf}.\maxP + \max(\maxc_i,0), \minc^*)$;\quad
                    $\maxc^* \gets \max(\maxC - \mathit{tpf}.\minP + \min(\minc_i,0), \maxc^*)$\;
                }
                $\mathit{tpf} \gets \mathit{tpf}.\mathit{next}$\;
            }
            \If(\tcp*[f]{Evaluate upper boundary point $\maxs_i$}){$\mind_i \geq 1 \wedge \mathit{tpf}.\mathit{time} = \maxs_i$}{
                $\minc^* \gets \min(\minC - \mathit{tpf}.\maxP + \max(\maxc_i,0), \minc^*)$;\quad $\maxc^* \gets \max(\maxC - \mathit{tpf}.\minP + \min(\minc_i,0), \maxc^*)$\;
            }
            \If(\tcp*[f]{Apply free-part consumption consistency (TC2)}){$\mind_i \geq 1$}{
                $\minc_i \gets \max(\minc^*, \minc_i)$;\quad $\maxc_i \gets \min(\maxc^*, \maxc_i)$ \label{algline:h}\;
            }
            $\maxd^* \gets \max(\maxe_i - s', \maxd^*)$;\quad
            $\maxd_i \gets \min(\maxd^*, \maxd_i)$ \tcp*{Apply duration consistency (TC3)}\label{algline:durationEnd}
        }
    }
    \caption{timetabling($T,\minC,\maxC$)}
    \label{algo:timetabling} 
\end{algorithm}

In order to compute the maximum duration of the task, we maintain the variables~$\maxd^*$ which contains the maximum duration found so far and $s'$ which corresponds to the earliest time at which the task can start and span without capacity violation until the current time point.
During the loop, $\maxd^*$ is updated if the difference between the current time point time and $s'$ is greater than its previous value.
If the task is detected as infeasible during a time point, $s'$ is set to the end of this time point (Line~\ref{algline:durationUpdate}).
Note that in order to take into account the parts of the time window that are not considered in Loop~\ref{algloop:ttfree}, $s'$ is initialized to $\mins$ and the difference between $\maxe$ and $s'$ is considered at the end of the loop (Line~\ref{algline:durationEnd}).

Similarly, when $\mind_i \geq 1$, the minimum and maximum available consumptions are computed as $\minc^*$ and $\maxc^*$ and used to update the consumption of the task if it has no fixed part (Line~\ref{algline:h}).
Note that these variables are initialized based on the consumption available at the time point before the start of Loop~\ref{algloop:ttfree} ($\mathit{tpf}.\mathit{prev}$) in order to consider the whole minimum overlapping interval of the task, and if a time point starts at $\maxs_i$, it is evaluated at the end of the scan to cover the upper boundary of the minimum overlapping interval; if $\mind_i = 0$, the task may execute with duration zero, so consumption is not restricted.

The \textit{Mandatory} rule is enforced over the time window outside the fixed part of each task.
The function \textit{checkIfMandatory} of Algorithm~\ref{algo:checkmandatory} is used to check this rule and apply its adjustments if needed.
For a required task ($i \in R$), this check is not needed over its fixed part $[\maxs_i,\mine_i)$ because the task is already established to execute across this interval, and its consumption is adjusted directly in Step~3.
If the task is optional ($i \in O$), its execution over $[\maxs_i, \mine_i)$ is not yet guaranteed, so \textit{checkIfMandatory} is evaluated at each time point in Step~3 at Line~\ref{algline:fpmand}.
Across all branches, \textit{checkIfMandatory} is called at Lines~\ref{algline:fwmand},~\ref{algline:bwmand},~\ref{algline:fpmand}, and~\ref{algline:freemand}.

\begin{algorithm}[tbp]
    \DontPrintSemicolon
    \If{
        $tp.\mathit{\#fp} > 0 \wedge (tp.\minP - \min(\minc_i, 0) > \maxC \vee \allowbreak tp.\maxP - \max(\maxc_i, 0) < \minC)$
    }{
        $p_i \gets \mathit{true}$ \;
        $\maxs_i \gets \min(tp.\mathit{time}, \maxs_i)$\;
        $\mine_i \gets \max(tp.\mathit{next}.\mathit{time}, \mine_i)$\;
        $\mathit{deficit} \gets \minC - (tp.\maxP - \max(\maxc_i, 0))$\; \label{deficit}
        $\mathit{overload} \gets \maxC - (tp.\minP - \min(\minc_i, 0))$\;\label{overload}
        \lIf{$\mathit{deficit} > 0$}{$\minc_i \gets \max(\mathit{deficit}, \minc_i)$}
        \lIf{$\mathit{overload} < 0$}{$\maxc_i \gets \min(\mathit{overload}, \maxc_i)$}
    }
    \caption{checkIfMandatory($i,tp,\minC,\maxC$)}
    \label{algo:checkmandatory} 
\end{algorithm}

Note that some optimizations are possible depending on the set of tasks given as input.
Indeed, the \textit{Mandatory} propagation rule is only useful if the tasks have a mix of positive and negative consumptions or if there is a minimum capacity~$\minC$ to enforce.
If all the consumption variables are fixed to a single value, the \textit{Consumption} adjustment rule is not necessary.
Similarly, the \textit{Duration} adjustment rule is only needed if the tasks have variable durations.
If these three rules can be ignored, only steps 1 and 2 need to be performed and the unnecessary loops at Lines~\ref{algloop:ttfp} and~\ref{algloop:ttfree} can be avoided.

Another optimization is to avoid considering some tasks when performing the timetabling algorithm.
Indeed, fixed tasks that occur before the start of the earliest unfixed task or after the end of the latest unfixed task are not useful for filtering.
Thus, such tasks do not need to be processed which reduces the number of time points in the profile.
To do so, we use the same process as the \emph{Fruitless Fixed Tasks Removal} of~\textcite{GayHS15}.

The following theorems establish the soundness, filtering consistency, and worst-case time complexity of Algorithm~\ref{algo:timetabling} (all proofs are deferred to Appendix~\ref{sec:proofs}).

\begin{theorem}[Soundness of Algorithm~\ref{algo:timetabling}]
\label{thm:tt-soundness}
Algorithm~\ref{algo:timetabling} is \emph{sound}: every domain reduction it performs removes only values that cannot appear in any solution of $\gcumulative(T,\minC,\maxC)$.
\end{theorem}

\begin{theorem}[Timetabling Filtering Consistency]
\label{thm:tt-consistency}
Upon termination without failure, Algorithm~\ref{algo:timetabling} achieves timetabling filtering consistency (Definition~\ref{def:tt-consistency}) with respect to the timeline~$\mathcal{L}$ computed by Algorithm~\ref{algo:init}.
\end{theorem}

\paragraph{Propagator Idempotency and Engine Fixpoint}
While Theorem~\ref{thm:tt-consistency} proves timetabling filtering consistency with respect to the fixed timeline~$\mathcal{L}$, Algorithm~\ref{algo:timetabling} is not idempotent because task domains are tightened in-place while $\mathcal{L}$ remains static. An immediate re-execution of Algorithm~\ref{algo:timetabling} on the narrowed domains could rebuild an updated timeline and achieve further pruning without external search branching.
Algorithm~\ref{algo:timetabling} intentionally performs a single filtering pass over $\mathcal{L}$ to guarantee an $\mathcal{O}(n^2)$ worst-case time bound per invocation.
Global fixpoint consistency is achieved through the standard propagation queue of the CP solver engine, which automatically re-enqueues the propagator whenever variable bounds are modified.

\begin{theorem}[Complexity of Algorithm~\ref{algo:timetabling}]
\label{thm:tt-complexity}
Algorithm~\ref{algo:timetabling} has a worst-case time complexity of~$\mathcal{O}(n^2)$, where $n = |T|$.
\end{theorem}

\begin{example}
When executing the timetabling algorithm on the same tasks as in Example~\ref{ex:profile}, the adjustments are:
\begin{enumerate}
    \item When processing task~$A$ at time point~$1$, its maximum consumption~$\maxc_A$ is adjusted to $1$ by the \textit{Consumption} rule.
    \item When processing task~$B$ at time point~$2$, the task is detected as conflicting at this time point by the \textit{Forbid} rule.
    Its minimum start $\mins_B$ is thus pushed to the next time point at time~$3$ where the adjustment stops as the task is no longer conflicting at this time.
    This also updates its minimum end to $\mine_B = 6$ due to the internal relation $\mine_B \geq \mins_B + \mind_B$.
    \item When processing task~$C$ at time point~$4$, the task is detected as mandatory.
    The following adjustments are done:
    \begin{enumerate}
        \item The task is set to present.
        \item Its maximum starting time is set to $\maxs_C = 4$ which also updates its maximum ending time to $\maxe_C = 7$ due to the internal relation $\maxe_C \leq \maxs_C + \maxd_C$.
        \item Its minimum ending time is set to $\mine_C = 5$.
        \item As the value $overload = -1$ is negative, the maximum consumption of the task is set to $\maxc_C = -1$.
    \end{enumerate}
    Note that even if task~$B$ is processed before task~$C$ and its fixed part increased, task~$C$ is adjusted based on the profile range which has been computed based on the state of task~$B$ at the start of the propagation.
\end{enumerate}
The state of the tasks after timetabling is:
\begin{description}
    \item[$A$] 
        : $\langle
        s = [0,1],\allowbreak
        d = [3,4],\allowbreak
        e = [3,4],\allowbreak
        c = 1,\allowbreak
        p = \mathit{true}
        \rangle$

    \item[$B$]
        : $\langle
        s = [3,4],\allowbreak
        d = [3,4],\allowbreak
        e = [6,7],\allowbreak
        c = 2,\allowbreak
        p = \mathit{true}
        \rangle$
    
    \item[$C$]
        : $\langle
        s = [3,4],\allowbreak
        d = [1,3],\allowbreak
        e = [5,7],\allowbreak
        c = [-2,-1],\allowbreak
        p = \mathit{true}
        \rangle$
\end{description}
Figure~\ref{fig:profileFiltered} shows the state of the tasks as well as the adjustments done.
    \begin{figure}[tbp]
        \centering
        \includegraphics[width=.7\linewidth]{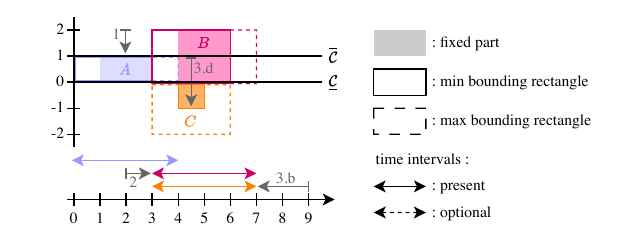}
        \caption{Representation of the tasks of Example~\ref{ex:profile} after timetabling. Adjustments are shown as gray arrows when applicable.}
        \label{fig:profileFiltered}
    \end{figure}
\end{example}

\section{Related Work}
\label{sec:rlw}

This section reviews existing approaches for modeling and solving producer-consumer constraints in constraint programming.
We organize it around three main categories: foundational work on producer-consumer constraints, generalized multi-resource constraints, and closed-source commercial solvers.

\subsection{Producer-Consumer constraints}

The foundational work on producer--consumer problems in constraint-based scheduling was established by~\textcite{cesta1997time} and extended by~\textcite{laborie2003algorithms}.
Laborie models resource constraints as tuples $\langle T, R, c, TE \rangle$, 
where $c$ denotes the quantity of the resource~$R$ consumed (if $c < 0$) or produced (if $c > 0$) by activity $T$, 
and $TE$ specifies the time extent during which the resource availability is affected. 
The proposed time extents include \textit{FromStartToEnd}, \textit{AfterStart}, \textit{AfterEnd}, \textit{BeforeStart}, \textit{BeforeEnd}, and \textit{Always}.
Notice that these constructs are conceptually close to cumulative functions constructs on conditional time intervals, although they do not explicitly capture the conditional nature of the latter.
The filtering descriptions in~\textcite{cesta1997time} and~\textcite{laborie2003algorithms} are quite different from ours as they both rely on temporal networks and precedence graphs between so-called production and consumption events, 
rather than on the profile-based reasoning over mandatory parts used in this paper.
Nevertheless, the filtering looks similar in spirit to the timetabling algorithm we present.
For instance, \textcite{cesta1997time} describes an optimistic resource profile and a pessimistic resource profile, 
which are similar to our minimum and maximum profiles. 
The proposed algorithm achieves a worst-case time complexity of~$\mathcal{O}(n^2)$, where $n$ is the number of tasks. 
However, it does not provide implementation details, such as the profile data structures.
The work of~\textcite{schutt2016explaining} builds on these ideas to provide explanations suitable to be used by lazy clause generation solvers.

\subsection{Generalized multi-resource constraint}\label{subsec:surplus}

\textcite{BeldiceanuC02} introduced a timetabling filtering algorithm for the \gcumulative\ constraint with $k$ alternative resources. 
This constraint does not rely on conditional task intervals or cumulative functions (which were only introduced later by~\textcite{LaborieRSV09}); instead, it operates directly on arrays of integer variables. 
The signature of the constraint is:  
$$cumulatives(s[1,n],d[1,n],e[1,n],c[1,n],m[1,n],\underline{\mathcal{C}}[1,k],\overline{\mathcal{C}}[1,k])$$  
where $s$, $d$, $e$, and $c$ denote respectively the start times, durations, end times, and consumptions (which can be positive or negative).  
The variables~$m$ indicate, for each task, the resource on which it executes, with domain on $\{1,\ldots,k\}$.  
Finally, $[\underline{\mathcal{C}}[j],\overline{\mathcal{C}}[j]]$ specifies the capacity range of resource~$j$.

Although this constraint does not directly support conditional task intervals, it is straightforward to connect the two modeling approaches.
Indeed, a dummy resource can be used to indicate if a task is not present in the multi-resource formulation.
Conversely, the conditional task approach supports multiple resources by using
an additional \emph{alternative} constraint~\cite{LaborieR08} that represent the choice of alternative resources.
This requires duplicating the activity once for each alternative resource, but it has the advantage that each alternative task can be filtered independently during propagation.  
In contrast, the multi-resource formulation propagation can only exploit the minimum over the different possible starts across all resources, which can lead to a weaker filtering.

The filtering algorithm of~\textcite{BeldiceanuC02} proceeds by collecting a set of tasks at relevant time points before performing propagation. This algorithm is implemented in the open-source solver \Gecode and according to its documentation, the same algorithm is also implemented in \SICStus. 
The algorithm presents several substantial differences with our algorithm, summarized in Table~\ref{tab:comparison-gecode}.

\begin{table}[tbp]
\centering
\small
\begin{tabular}{@{}p{3.2cm}p{5.3cm}p{5.5cm}@{}}
\hline
\textbf{Feature / Aspect} & \textbf{\textcite{BeldiceanuC02}} & \textbf{This Work} \\ \hline
Data structure & Event array sweep & Doubly-linked \textit{Profile} with $\mathcal{O}(1)$ boundary pointers \\
Backward adjustment & Single adjustment per propagation call & Chained adjustments in a single pass (Ex.~\ref{ex:Lct}) \\
Consumption pruning & Restricted to single-event span & Minimum overlapping interval $[\mine_i-1, \maxs_i]$ (Ex.~\ref{ex:height}) \\
Duration pruning & Measured from single conflicting point & Longest conflict-free sub-interval (Ex.~\ref{ex:duration}) \\
Optional tasks & Indirect via dummy resource & Native conditional intervals with direct presence pruning \\ \hline
\end{tabular}
\caption{Key differences between the timetabling algorithm of~\textcite{BeldiceanuC02} and the profile-based algorithm proposed in this work.}
\label{tab:comparison-gecode}
\end{table}

First, our approach builds on the profile-based paradigm of~\textcite{GayHS15,GingrasQ16} for the classical cumulative constraint, which constructs an explicit resource profile and tests each task against it.  
We rely on the doubly-linked \textit{Profile} data structure of~\textcite{GingrasQ16} to prune task bounds both forward and backward, retrieving the starting rectangle in~$\mathcal{O}(1)$ via direct pointers.  
In contrast to~\textcite{GayHS15}, which represents the profile as an unsorted list of rectangles, direct boundary pointers eliminate search overhead.  
Consequently, although the worst-case quadratic time complexity per call remains $\mathcal{O}(n^2)$ as in~\textcite{BeldiceanuC02}, our algorithm performs substantially fewer fixpoint iterations and avoids redundant passes.  
Second, as demonstrated in the following examples, our algorithm prunes the consumption and duration attributes of tasks more effectively than~\textcite{BeldiceanuC02}.

\paragraph{Backward Propagation Difference}
For the adjustment of the maximum end time ($\maxe$) of tasks due to the \textit{Forbid} rule, when a task requires two or more consecutive adjustments, our algorithm performs all the adjustments in a single call to the timetabling algorithm compared to the version of~\textcite{BeldiceanuC02} where only one adjustment is done per call.
Indeed, in~\textcite{BeldiceanuC02}, detection and adjustment are performed only in a forward fashion.
Thus, the filtering of the maximum end time of tasks is done only in regards to the current end of the task time window.
This means that if a task has its maximum end time adjusted, whether or not the maximum end time can be further adjusted will be considered in the next call to the filtering algorithm.
In practice, this may lead to situations where the filtering algorithm of~\textcite{BeldiceanuC02} needs several consecutive calls in order to completely adjust the end time of a task and reach a fixed point.
In contrast the algorithm presented in this paper does such adjustments in a single pass as illustrated in Example~\ref{ex:Lct}.
\begin{example}
\label{ex:Lct}
Let us consider three tasks:
\begin{itemize}
    \item Task A is fixed and present, starts at 4, ends at 5 and has a consumption of 1.
    \item Task B is fixed and present, starts at 7, ends at 8 and has a consumption of 1.
    \item Task C is present, has a minimum start of 0, a maximum end of 10, a fixed duration of 3 and a consumption of 1.
\end{itemize}
The maximum capacity of the resource~$\maxC$ is 1.
Figure~\ref{fig:exampleLct} shows the three tasks before propagation.
The dashed blue rectangle represents the time window $[\mins, \maxe)$ of task C.
\begin{figure}[tbp]
    \centering
    \includegraphics[width=0.6\linewidth]{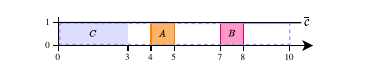}
    \caption{Example~\ref{ex:Lct}: three tasks before timetabling with $\maxC=1$. Tasks~A ($[4,5)$, consumption~1) and B ($[7,8)$, consumption~1) are fixed and present; the dashed blue rectangle shows the time window $[\mins_C, \maxe_C) = [0,10)$ of task~C (fixed duration~3, consumption~1).}
    \label{fig:exampleLct}
\end{figure}

During the backward propagation of our timetabling algorithm, task C is detected as infeasible at times 7 then 4, and its maximum end time is successively updated to 7 then 4 in a single call.
The propagation from~\textcite{BeldiceanuC02} iterates the profile in one direction and detects the infeasibility of task C only at time 7 and adjusts its maximum end at 7 during the first call.
A second call to the algorithm is necessary in order to adjust the maximum end at time 4.
\end{example}

\paragraph{Consumption Adjustment Difference}

For consumption adjustment, the difference occurs in the case where a task has no fixed part (second case of the \textit{Consumption} filtering rule).
In this case, our algorithm considers the maximum consumption available over the minimum overlapping interval of the task~\cite{gay2015time}.

The propagation from~\textcite{BeldiceanuC02} uses another consumption adjustment rule (see Alg. 4 in their paper).
It considers the maximum available consumption only if the task examined has both its earliest completion time ($\mine$) and its latest start time ($\maxs$) in the same event interval.
That means that if one or more other tasks affect the available consumption between the $\mine$ and $\maxs$ of a task, leading them to occur during different events, the consumption of the task is not adjusted.
Example~\ref{ex:height} illustrates this.

\begin{example}
\label{ex:height}
Let us consider three tasks:
\begin{itemize}
    \item Task A is fixed and present, starts at 4, ends at 12 and has a consumption of 2.
    \item Task B is fixed and present, starts at 6, ends at 10 and has a consumption of -1.
    \item Task C has a minimum start of 0, a maximum end of 16 and a fixed duration of 6.
    Its consumption is in the interval $[1, 4]$
\end{itemize}
The maximum capacity of the resource~$\maxC$ is 4.
Figure~\ref{fig:exampleHeight} shows the three tasks before propagation.
The dashed blue rectangle represents task C maximum bounding rectangle.
The dotted black line shows the minimum profile $\minP$.
\begin{figure}[tbp]
    \centering
    \includegraphics[width=0.6\linewidth]{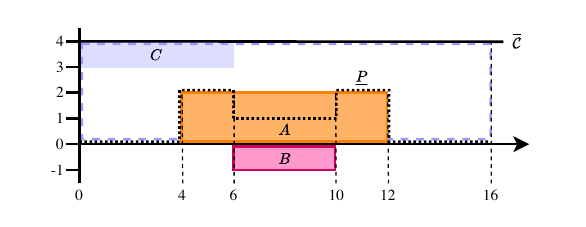}
    \caption{Example~\ref{ex:height}: three tasks before timetabling with $\maxC=4$. Tasks~A ($[4,12)$, consumption~2) and B ($[6,10)$, consumption~$-1$) are fixed; the dashed blue rectangle shows task~C's maximum bounding rectangle (time window~$[0,16)$, fixed duration~6, consumption~$\in[1,4]$). The dotted black line is the minimum profile~$\minP$.}
    \label{fig:exampleHeight}
\end{figure}

During the execution of the timetabling algorithm, the maximum consumption of task C is adjusted to 3 as this value corresponds to the maximum available consumption over its minimum overlapping interval (from time 5 to time 10).
This filtering does not occur with the \gcumulative\ constraint from~\textcite{BeldiceanuC02} as the $\mine_C$ and $\maxs_C$ of task C are not in the same event due to the presence of task B.
If the task B is removed, then both algorithms obtain the same filtering as the consumption of task C is adjusted to 2.
\end{example}

\paragraph{Duration Adjustment Difference}

The difference in filtering on the maximum duration adjustment of the tasks occurs
if the profile prevents a task from being present in at least two different parts of its time window.
In this case, the algorithm presented in this paper identifies the longest interval in the profile where the task can fit and adjusts the task's maximum duration accordingly.

In contrast, the filtering of~\textcite{BeldiceanuC02} considers one single conflict point at a time, rather than the entire conflicting interval.
When encountering conflicting time point in the profile, the task maximum duration is updated based on the maximum span between the conflicting time point and either the minimum start or the maximum end of the task.
Example~\ref{ex:duration} illustrates this.

\begin{example}
\label{ex:duration}
Let us consider three tasks:
\begin{itemize}
    \item Task A is fixed and present, starts at 3, ends at 6 and has a consumption of 3.
    \item Task B is fixed and present, starts at 10, ends at 14 and has a consumption of 3.
    \item Task C has a minimum start of 0, a maximum end of 16 and a consumption of 2.
    Its duration is in the interval $[2, 16]$
\end{itemize}
The maximum capacity of the resource~$\maxC$ is 4.
Figure~\ref{fig:exampleDuration} shows the three tasks before propagation.
The dashed blue rectangle represents task C maximum bounding rectangle.
The black dotted line shows the maximum capacity ($\maxC$).
In this example, both tasks A and B prevent task C from being present at the same time.
\begin{figure}[tbp]
    \centering
    \includegraphics[width=0.6\linewidth]{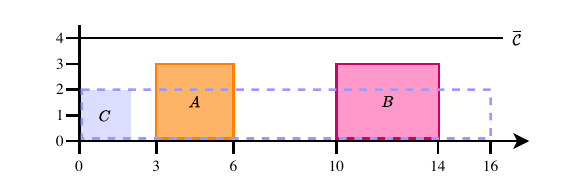}
    \caption{Example~\ref{ex:duration}: three tasks before timetabling with $\maxC=4$. Tasks~A ($[3,6)$, consumption~3) and B ($[10,14)$, consumption~3) are fixed; the dashed blue rectangle shows task~C's maximum bounding rectangle (time window~$[0,16)$, consumption~2, duration~$\in[2,16]$). The dotted black line shows the maximum capacity~$\maxC$.}
    \label{fig:exampleDuration}
\end{figure}

During the execution of the timetabling algorithm, the maximum duration of task C is adjusted to 4, as this is the span of the longest feasible placement interval (from time 6 to time 10).
When filtering with the propagation algorithm from~\textcite{BeldiceanuC02}, the duration is instead adjusted to 10, since this corresponds to both the span available after task A and the span available before task B.
\end{example}

\subsection{Closed source solvers}

Both \CPOptimizer and \OptalCP natively implement the cumulative functions modeling paradigm.
While the conceptual foundations of conditional intervals and cumulative function algebra have been extensively documented by~\textcite{LaborieR08,LaborieRSV09,laborie2018ibm}, both engines are closed-source commercial software.
Consequently, their concrete algorithmic implementations, internal profile data structures, and propagation mechanics for range constraints on cumulative functions have not been published.
Based on the filtering and the number of backtracks on some examples, we observed that \CPOptimizer achieves a filtering similar to the \Gecode implementation when a cumulative function requires a generalized cumulative constraint.
For specific cases, like when a cumulative function models a non-overlap constraint, the constraint is reformulated by \CPOptimizer into a standard disjunctive constraint for which strong dedicated filtering exists (such as edge-finding, not-first/not-last, etc.).

Contrary to ours, the \CPOptimizer modeling API does not allow cumulative functions to take negative values.
Indeed, while cumulative tasks can have a negative contribution, a cumulative function whose value becomes negative will cause a failure in \CPOptimizer.
This is equivalent to having a constraint $f \geq 0$ on any cumulative function~$f$ used in the model.
It is not known whether this limitation is due to a design choice or a limitation in the filtering algorithm used for the constraint.

\section{Experiments}\label{sec:xp}

Our approach and filtering algorithm are open source and available in \MaxiCP, a Java solver that extends \MiniCP.
We compare it with the closed-source state-of-the-art commercial solver \CPOptimizer as well as the open-source solver \Gecode that implements a version of the \gcumulative\ constraint of~\textcite{BeldiceanuC02}.
As cumulative functions are not available in \Gecode, we carefully modeled them in a way similar to how they would be flattened by Algorithm~\ref{algo:flatten}.
Three cumulative problems that involve reservoirs or negative consumptions are considered.
We use the same fixed static search strategy for all solvers and problems to ensure that only the filtering strength and propagation speed influence the results.
Although \OptalCP also supports cumulative functions, it does not allow fixing the search strategy as \CPOptimizer does.
Since this paper focuses on filtering and modeling rather than overall solving performance, including a solver with a free search would conflate search heuristics with propagation strength, making it impossible to isolate the contribution of the filtering algorithm; we therefore exclude it from the comparison.
The experiments were carried out on a MacBook Pro M3 with 32~GB of memory.
The source code of the models is available at~\cite{generalizedcumulativeSource}.

For each problem, we report results using a cactus plot showing the cumulative number of instances solved as a function of time.
We also provide a pairwise comparison of solvers in terms of the number of backtracks on instances solved by both solvers.

Each model involves a set of tasks~$i \in T$. Recall that the attributes for each task are
$s_i$ the start time, $d_i$ the duration, $e_i$ the end time, $c_i$ the consumption, $p_i$ the execution status (present or absent) when applicable.

\subsection{The RCPSP with Consumption and Production of Resources (RCPSP-CPR)} 
The Resource-Constrained Project Scheduling Problem with Consumption and Production of Resources (RCPSP-CPR)~\cite{kone2013comparison} is an extension of the classic Resource-Constrained Project Scheduling Problem (RCPSP) \cite{Dike_RCPSP,DING_2023_RCPSP} with additional reservoir resources that must always be kept positive.
Each task is present and, in addition to classical renewable resource consumption (modeled with pulse functions), tasks also consume a specified amount of reservoir resources at their start and produce a specified amount at their end (modeled with stepAtStart and stepAtEnd).
The objective is to minimize the makespan.

\paragraph{Model}
$T$ denotes the set of tasks, $Rn$ is the set of renewable resources, $Rs$ is the set of reservoir resources, and $P$ is the set of precedences between pairs of tasks.
The capacity of each renewable resource~$k \in Rn$ is denoted $\mathcal{C}_k$ and the initial level of each reservoir resource~$u \in Rs$ is denoted $C^r_u$.
Each task~$i$ consumes $c_{ki}$ units of the renewable resource~$k \in Rn$, consumes $c^-_{ui}$ units of reservoir resource~$u\in Rs$ at its start time and produces $c^+_{ui}$ units of reservoir resource~$u\in Rs$ at its end time.

The model is written as:
\begin{align}
    & \mathit{minimize} \; \max\limits_{i\in T} e_i & \label{obj} \\
    & \text{subject to} & \nonumber \\
    & e_i \leq s_j & \forall (i,j)\in P \label{preced}\\
    & ren_k = \sum\limits_{i\in T} pulse(i, c_{ki}) & \forall k\in Rn \label{setRenew} \\
    & res_u = 
    \mathit{step}(0, C^r_u) - \sum\limits_{i\in T} \mathit{stepAtStart}(i, c^-_{ui}) + \sum\limits_{i\in T} \mathit{stepAtEnd}(i, c^+_{ui}) & \forall u\in Rs \label{setReser}\\
    & \mathit{alwaysIn}(ren_k, -\infty, \mathcal{C}_k) & \forall k\in Rn \label{postRenew}\\
    & \mathit{alwaysIn}(res_u, 0, +\infty) & \forall u\in Rs \label{postReser} 
\end{align}
The objective \eqref{obj} is to minimize the makespan.
The constraints \eqref{preced} ensure the precedences between tasks.
The renewable and reservoir resources are constrained in~\eqref{setRenew} and~\eqref{setReser} respectively. 
The resource constraints are posted at \eqref{postRenew} and~\eqref{postReser} respectively.

The search consists in a static binary branching that selects variables in the order in which tasks are declared in the instance file.
The first un-assigned start variable is selected and assigned to its minimum value in the left branch.
The right branch removes the minimum value from the domain.

\begin{figure}[tbp]
    \centering
    \includegraphics[width=0.72\linewidth]{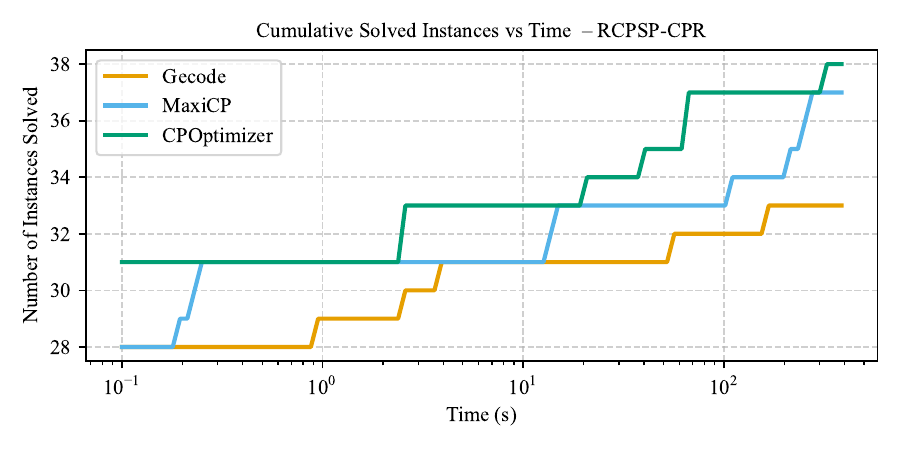}
    \caption{Number of solved instances as a function of time for the RCPSP-CPR Problem.}
    \label{fig:res_RCPSP}
\end{figure}

\paragraph{Results}
We experiment on 55~instances with 10~tasks adapted from~\textcite{kone2013comparison}.
A time limit of 600~s is set per instance for finding the first feasible solution.
As shown in Figure~\ref{fig:res_RCPSP}, \MaxiCP solves more instances than \Gecode and roughly the same set of instances as \CPOptimizer.

\begin{figure}[tbp]
    \centering
    \includegraphics[width=0.33\linewidth]{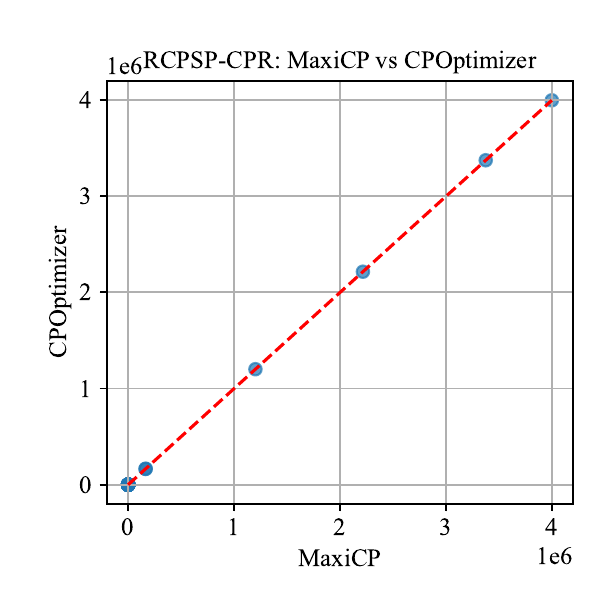}\hfill
    \includegraphics[width=0.33\linewidth]{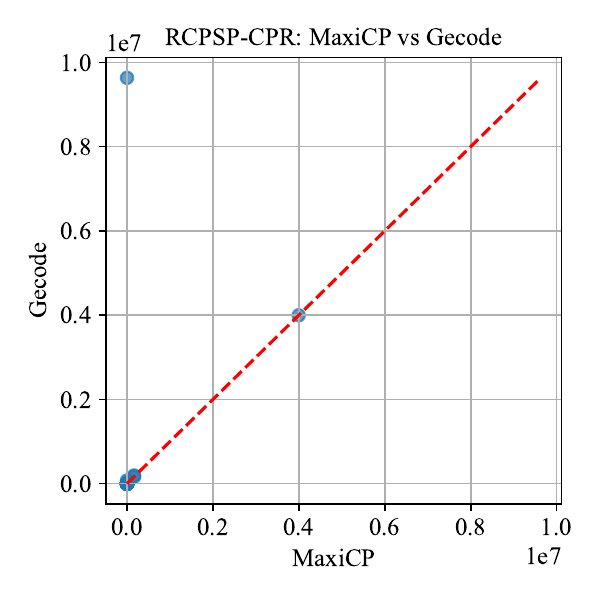}\hfill
    \includegraphics[width=0.33\linewidth]{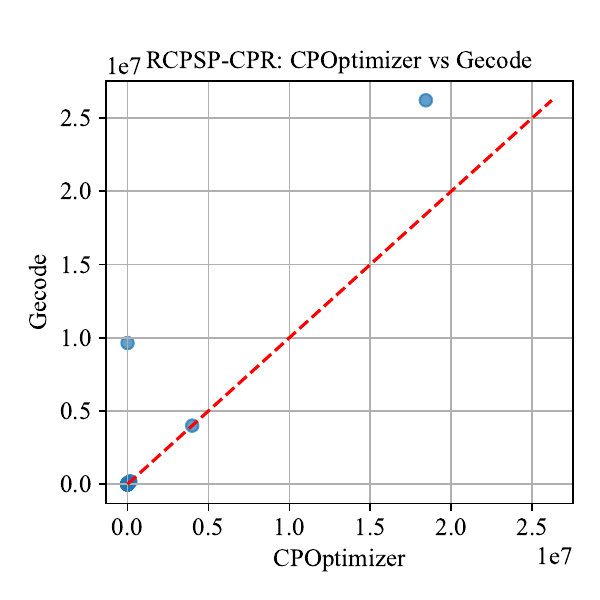}
    \caption{Failure comparison for the RCPSP-CPR}
    \label{fig:back_rcpsp}
\end{figure}

Figure~\ref{fig:back_rcpsp} reports for each pair of solvers, for each instance solved commonly by both solvers, an $x,y$-plot of the number of backtracks.
The comparison between \CPOptimizer and \MaxiCP reveals that the number of backtracks is nearly identical, suggesting that \CPOptimizer is simply faster on this problem, 
as both solvers achieve similar levels of filtering on this problem, but \MaxiCP requires fewer backtracks than \Gecode on some of the instances.

\subsection{The Single Machine with Inventory Constraints (SMIC)} 
The Single Machine with Inventory Constraints (SMIC) problem was introduced by~\textcite{smic}.
Each task has a release date $release_i$, a fixed duration, and a positive or negative inventory consumption~$invent_i$ that occurs at the start of the activity (modeled with stepAtStart).
The inventory starts at a given value~$\mathit{initInventory}$ and must stay within a specified range ($[0, \mathit{capaInventory}]$).
Activities cannot overlap in time.
The objective is to minimize the makespan.

\paragraph{Model}
The model is written as:
\begin{align}
& \mathit{minimize} \;  \max_{i \in T} e_i & \label{obj_smic} \\
& \text{subject to} & \notag \\
& s_i \geq \mathit{release_i}, & \forall i \in T \label{release} \\
& res = \mathit{step}(0, \mathit{initInventory}) + \sum_{i \in T} \mathit{stepAtStart}(i, invent_i) & \label{reservoir} \\
& \mathit{alwaysIn}(res, 0, \mathit{capaInventory}) & \label{reservoirConst} \\
& \mathit{noOverlap} = \sum_{i \in T} pulse(i, 1) & \label{noOverlap} \\
& \mathit{alwaysIn}(\mathit{noOverlap}, -\infty, 1) & \label{noOverlapConst}
\end{align}

The objective \eqref{obj_smic} is to minimize the makespan (i.e., the time at which all tasks are completed).
The release dates are ensured by the relation \eqref{release}.
The reservoir resource is constrained in relations \eqref{reservoir} and~\eqref{reservoirConst} to ensure that the capacity of the inventory is not violated.
The noOverlap constraint is enforced with the relations \eqref{noOverlap} and~\eqref{noOverlapConst} to ensure the no-overlapping of activities. Notice that those are modeled using cumulative functions rather than with global dedicated non-overlap constraints on task intervals to ensure that the results are solely impacted by the filtering of the cumulative constraint rather than by the possibly different filtering of the non-overlap constraint.
This, unfortunately, did not appear possible for \CPOptimizer which recognizes that the cumulative function is used to model a non-overlap constraint and automatically reformulates it using a non-overlap constraint with stronger filtering.
To avoid this bias, we use the timetabling filtering algorithm adapted from this paper for \CPOptimizer, to ensure consistent comparison with other solvers.

The search consists in a static branching heuristic that selects variables in the order in which tasks are declared in the instance file.
The first un-assigned start variable is selected and assigned to its minimum value in the left branch.
The right branch removes the minimum value from the domain.

\begin{figure}[tbp]
    \centering
    \includegraphics[width=0.72\linewidth]{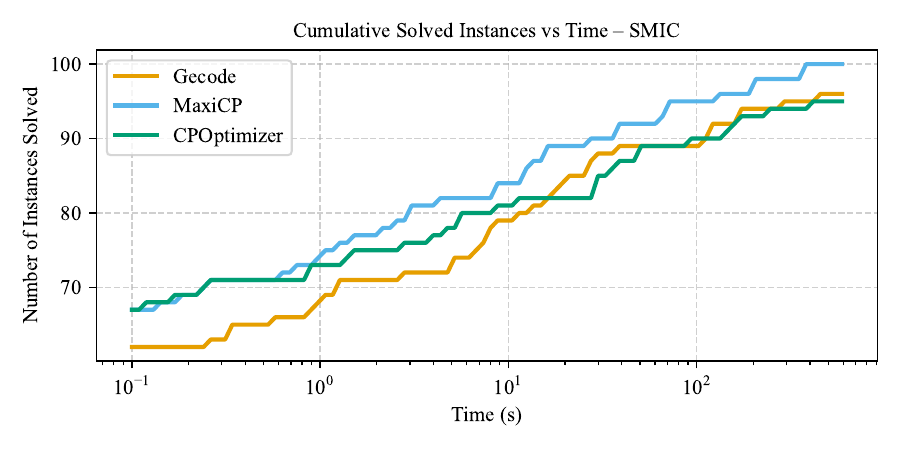}
    \caption{Number of solved instances as a function of time for the SMIC Problem}
    \label{fig:res_SMIC}
\end{figure}

\paragraph{Results}
We experiment on instances with 10 and 20~tasks from~\textcite{smic}.
A time limit of 600~s is set per instance for finding and proving optimality.
As shown in Figure~\ref{fig:res_SMIC}, \CPOptimizer solves fewer instances while \MaxiCP and \Gecode roughly have the same behavior on this problem.

\begin{figure}[tbp]
    \centering
    \includegraphics[width=0.33\linewidth]{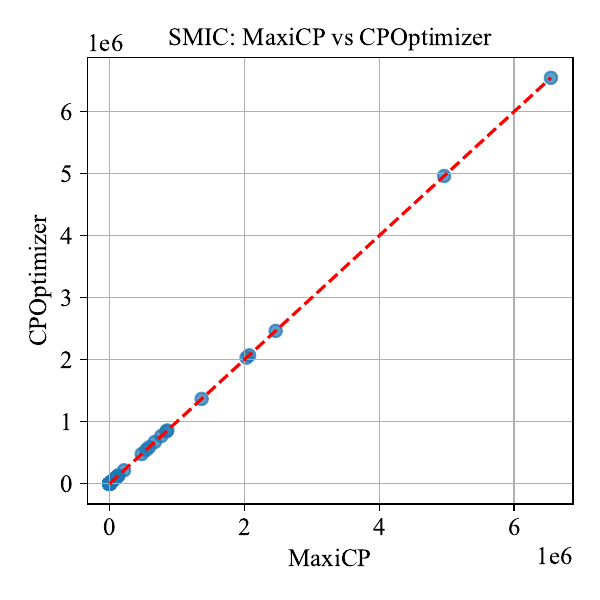}\hfill
    \includegraphics[width=0.33\linewidth]{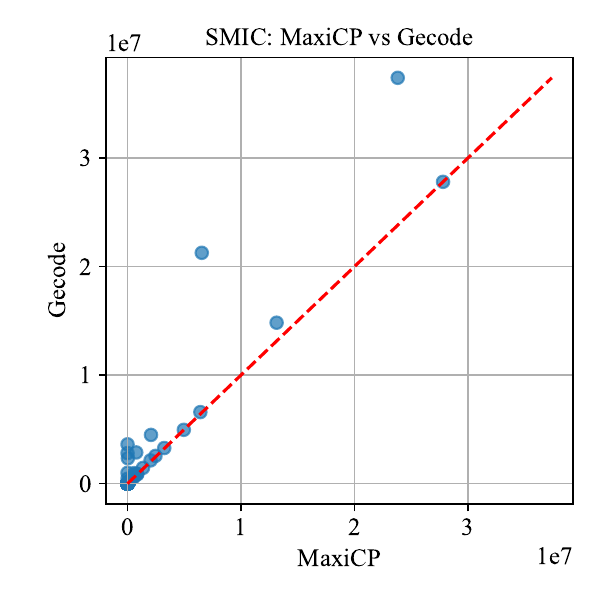}\hfill
    \includegraphics[width=0.33\linewidth]{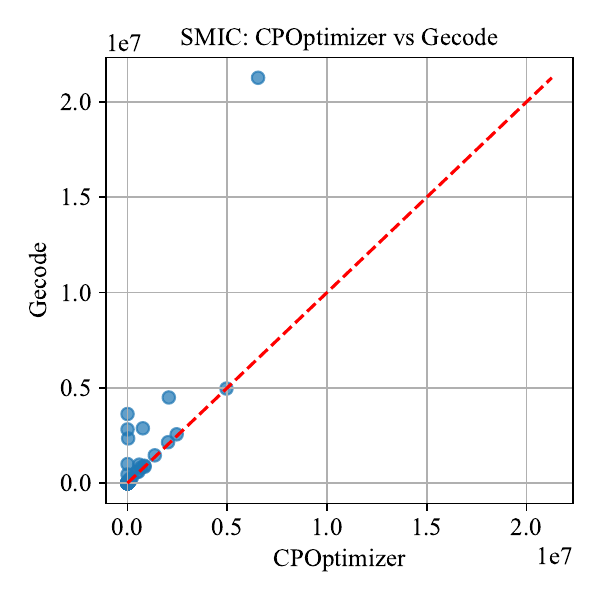}
    \caption{Failure comparison for the SMIC}
    \label{fig:back_smic}
\end{figure}

Figure~\ref{fig:back_smic} reports for each pair of solvers, for each instance solved commonly by both solvers, an $x,y$-plot of the number of backtracks.
This analysis shows that \CPOptimizer and \MaxiCP require the same number of backtracks on this problem, 
confirming comparable filtering strength between the two solvers. \Gecode, in contrast, misses some of the adjustments discussed in the sub-section~\ref{subsec:surplus},
resulting in weaker pruning; nonetheless, the number of instances it solves remains close to that of \CPOptimizer.

\subsection{The Maximum Energy Scheduling Problem (MESP)}
The Maximum Energy Scheduling Problem (MESP) consists in scheduling optional tasks with variable durations and resource demands, which can be positive or negative, on a single resource with fixed capacity~$\maxC$.
The energy of a task is the product of its duration and its demand.
The objective is to maximize the total positive energy consumption of the resource while ensuring that the resource capacity is not exceeded at any time.
All filtering rules are beneficial for this problem, making it the most relevant benchmark for evaluating the proposed filtering algorithm.

\paragraph{Model}
The model is written as:
\begin{align}
    & \mathit{maximize} \; \sum\limits_{i \in T \mid p_i = \mathit{true}} \max(c_i,0) \cdot d_i  \label{max_energy}\\
    & \text{subject to} \nonumber \\
    & \mathit{resource} = \sum\limits_{i\in T} \mathit{pulse}(i, c_{i}) \label{setResMESP} \\
    & \mathit{alwaysIn}(\mathit{resource}, -\infty, \maxC)  \label{postResMESP}
\end{align}
The objective \eqref{max_energy} is to maximize the sum of the positive energies of the tasks that are present.
The constraint~\eqref{setResMESP} sets up the reservoir resource and its capacity is constrained by~\eqref{postResMESP}.
While MESP is formulated as an optimization problem in~\eqref{max_energy}, our scalability experiment evaluates the performance of the filtering algorithm when driven by a greedy constructive heuristic that seeks to maximize energy in a single dive without backtracking, stopping at the first feasible solution.
This isolates propagation throughput on massive task sets from search exploration. 

Note that the resource demand variables of tasks~$c_i$ may have negative values in their domain in this problem.
As explained in \textit{Filtering Differences} section, \CPOptimizer does not allow cumulative functions to have a negative value.
In order to accommodate this restriction, the model for \CPOptimizer has been adapted in the following way:
The cumulative function~$resource$ is shifted up in height at the time 0 by a step function of height $neg^* = \sum_{i \in T} -\min(\minc_i, 0)$ which corresponds to the sum of all maximum negative values of the tasks' resource demands.
The capacity~$\maxC$ of the resource is also shifted by the same value.
In formal terms, constraint~\eqref{setResMESP} is changed to: $resource = step(0, neg^*) + \sum\limits_{i\in T} pulse(i, c_{i})$ and constraint~\eqref{postResMESP} is changed to $\mathit{alwaysIn}(\mathit{resource}, -\infty, \maxC + neg^*)$.
This ensures that the cumul function~$resource$ remains non-negative, as required by \CPOptimizer.

The search heuristic selects tasks in the fixed order of their declaration in the instance file.
For each task, variables are fixed greedily to their maximum domain values in the following order: presence $p_i \gets \mathit{true}$, demand~$c_i \gets \maxc_i$, duration~$d_i \gets \maxd_i$, and end time~$e_i \gets \maxe_i$.
Two branches are created: the left branch fixes the value while the right branch removes it from the domain.
The search stops at the first solution found.

\paragraph{Instance Generation}
We generated 60 MESP instances with the total number of tasks~$n$ ranging from 6 to 12{,}800.
For each specified size, five distinct instances were generated using different random seeds.
In every instance, the task set is divided into tasks drawing resource capacity (80\% of the total, i.e.\ $n - \lfloor n/5 \rfloor$ tasks) and tasks replenishing capacity (20\%, i.e.\ $\lfloor n/5 \rfloor$ tasks).
All instances share a fixed resource capacity of $\maxC = 10$ and a maximum task duration of $L_{\max} = 20$.
The scheduling horizon $H$ is set to $H = \max(50, 5n)$.
For tasks drawing resource capacity, the maximum resource demand~$\maxc_i$ is drawn uniformly from $[1, 10]$, and the minimum demand~$\minc_i$ from $[1, \maxc_i]$.
For tasks replenishing capacity, the maximum resource demand~$\maxc_i$ is drawn from $[-10, -1]$ and the minimum demand~$\minc_i$ from $[\maxc_i, -1]$.
The minimum start time of every task is drawn uniformly from $[0, H - L_{\max} - 1]$, ensuring no task is forced to extend beyond the horizon.

\begin{figure}[tbp]
    \centering
    \includegraphics[width=0.72\linewidth]{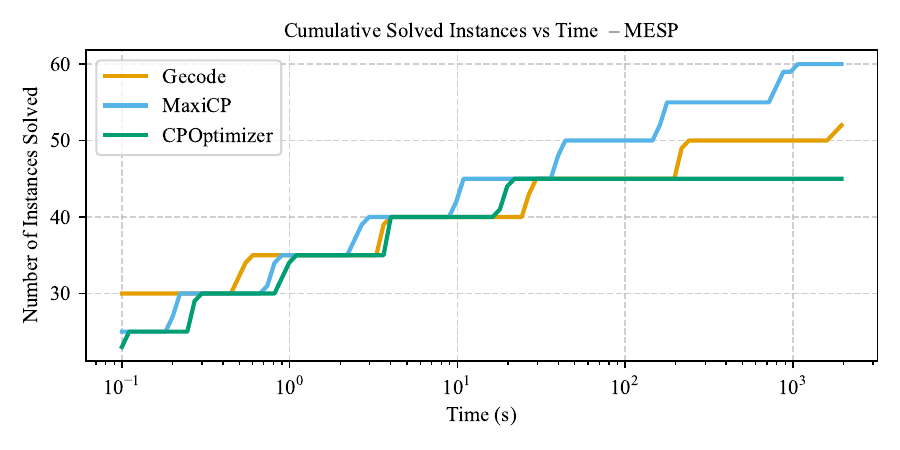}
    \caption{Number of solved instances as a function of time for the MESP Problem.}
    \label{fig:res_MESP}
\end{figure}

\paragraph{Results}
A time limit of 2,000~s and a memory limit of 16~GB is set per instance for finding a feasible solution.
Given the large number of tasks in some instances, this experiment is designed to test the scalability of the filtering.
We chose MESP for this scalability experiment because, unlike RCPSP-CPR and SMIC, its benchmark instances span a much wider 
range of sizes, with some instances containing several thousand tasks and resource requirements far exceeding those found in 
the other two problems. This large variability in instance size makes MESP particularly well-suited to stress-test the 
filtering algorithm under realistic, large-scale conditions, and to observe how its runtime and pruning behavior evolve as the 
problem size grows. In contrast, RCPSP-CPR and SMIC instances remain comparatively small and homogeneous in size, and would not 
allow us to draw meaningful conclusions about scalability.
As can be observed in Figure~\ref{fig:res_MESP}, our approach performs very well with more instances solved.
Both \Gecode and \CPOptimizer fail to solve the largest instances due to timeouts and memory limits respectively.

\begin{figure}[tbp]
    \centering
    \includegraphics[width=0.33\linewidth]{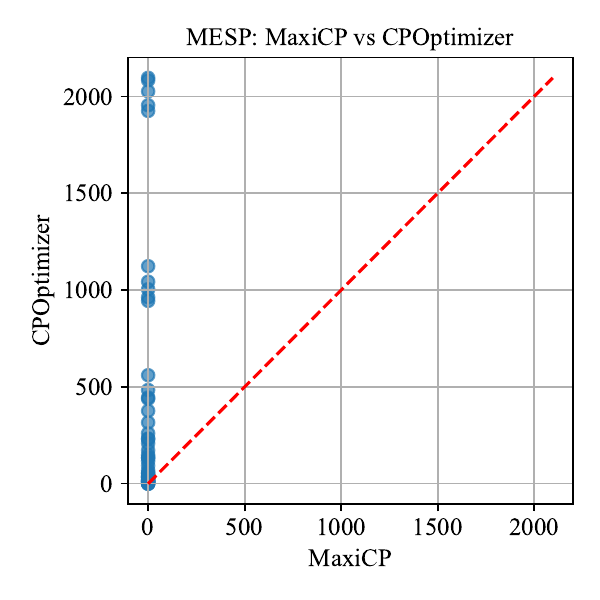}\hfill
    \includegraphics[width=0.33\linewidth]{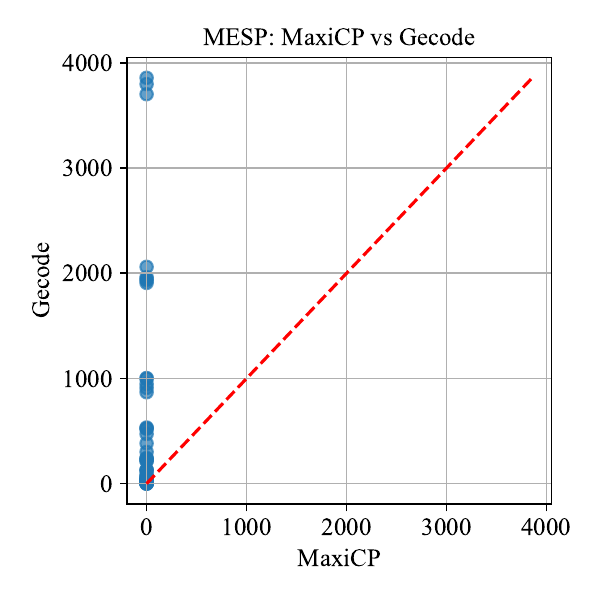}\hfill
    \includegraphics[width=0.33\linewidth]{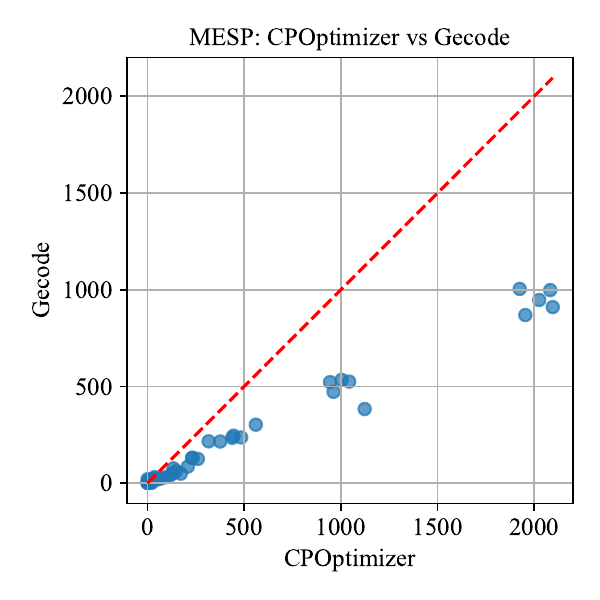}
    \caption{Failure comparison for the MESP}
    \label{fig:back_mesp}
\end{figure}

Figure~\ref{fig:back_mesp} reports for each pair of solvers, for each instance solved commonly by both solvers, an $x,y$-plot of the number of backtracks.
As can be seen, our approach avoids backtracking, unlike the other solvers.
Interestingly, \CPOptimizer requires fewer backtracks than \Gecode although it solves fewer instances.

\subsection{Why not compare with solvers using \MiniZinc?}
\label{subsec_why_not_minizinc}
The \MiniZinc modeling language provides a \texttt{cumulatives} constraint that allows for negative consumption. Unfortunately, with the exception of \SICStus and \Gecode, this constraint is decomposed in other solvers.
Also the \MiniZinc API for this constraint does not support explicit access to the \emph{end} variables. Being able to access the \emph{end} variables directly rather than relying on the relation \emph{start + duration = end} is crucial for the problems considered, as the \emph{end} variables can be constrained independently, as explained in the next example.

\begin{example}
Assume a task with the following properties: $s=[3,7]$, $d=[1,5]$, $e=[8,8]$. 
The mandatory part of this task is $[\overline{s},\underline{e}-1]=[7,7]$. 
If the end variable is not available, the mandatory part inferred from $s$ and $d$ is $[\overline{s},\underline{s}+\underline{d}-1]=\emptyset$.
\end{example}

As a result, the propagation when using \MiniZinc for these problems can be much weaker, making the comparison unfair.
This was confirmed experimentally and can be verified with \MiniZinc models.

\section{Conclusion}\label{sec:concl}

We presented an implementation of the cumulative functions modeling paradigm for scheduling problems.
Our implementation applies a flattening procedure to cumulative function expressions, producing a set of cumulative tasks with positive or negative resource consumption, which are then passed to a \gcumulative\ constraint.

We introduced a novel timetabling filtering algorithm for the \gcumulative\ constraint that operates on conditional time-interval variables and supports negative resource consumption.
This algorithm is simpler and achieves stronger pruning than the one proposed by~\textcite{BeldiceanuC02}.

Experimental results demonstrate that the proposed approach is scalable and competitive with other solvers on three cumulative scheduling problems with diverse characteristics, including optional tasks and negative resource consumption.

As future work, we plan to extend the filtering algorithm to generalize the Overload/Underload checking and edge-finding rules as in \cite{KameugneFND24,KameugneFND23,OuelletQ13} in the context of producers and consumers. We also intend to explore explanations for the generalized cumulative constraint, similar to those proposed by \textcite{schutt2011explaining}, and to investigate whether a CP-SAT solver such as OR-Tools~\cite{perron_et_al} could benefit from them.

\printbibliography

\appendix

\section{Proofs of Lemma and Theorems}\label{sec:proofs}

This appendix provides the formal proofs for all lemma and theorems presented in Section~\ref{sec:timetabling}.

\subsection{Proof of Lemma~\ref{lem:profile-bounds}}
\begin{proof}[Proof of Lemma~\ref{lem:profile-bounds}]
We prove the invariant maintained by the algorithm during the chronological processing of events.
At any moment, the running accumulators $\minP$ and $\maxP$ represent the contributions of all tasks whose possible time window overlaps the current time point, while $\mathit{\#fp}$ counts the required tasks whose fixed part overlaps the current time point.

Consider an arbitrary non-absent task~$i \in O \cup R$. Its possible execution window starts at $\mins_i$ and ends at $\maxe_i$. Therefore, the algorithm adds $\min(\minc_i,0)$ to $\minP$ and $\max(\maxc_i,0)$ to $\maxP$ at the event~$\mins_i$ and removes the same quantities at the event~$\maxe_i$. As a result, the task contributes to the accumulators exactly for those time points satisfying $\mins_i \leq tp.\mathit{time} < \maxe_i$.

Now suppose that $i$ is required and has a non-empty fixed part, i.e., $\maxs_i < \mine_i$. Its fixed part is active on the interval $[\maxs_i, \mine_i)$. The algorithm therefore adds $\max(\minc_i,0)$ to $\minP$ and $\min(\maxc_i,0)$ to $\maxP$ when processing $\maxs_i$, increments $\mathit{\#fp}$, and performs the symmetric removal and decrement at $\mine_i$. Hence a required task contributes to the fixed-part terms exactly when $\maxs_i \leq tp.\mathit{time} < \mine_i$, and the counter $\mathit{\#fp}$ is equal to the cardinality of this set.

Because the events are processed in increasing time order, each task is inserted and removed at the correct boundary times, and no contribution is missed or counted twice. Consequently, after all events with time at most $tp.\mathit{time}$ have been processed, the accumulators and the fixed-part counter coincide with the sums in~\eqref{eq:minP}--\eqref{eq:maxP} and the cardinality in the lemma. This proves the claim.
\end{proof}

\subsection{Proof of Theorem~\ref{thm:infeasibility}}
\begin{proof}[Proof of Theorem~\ref{thm:infeasibility}]
Suppose Algorithm~\ref{algo:init} returns \texttt{Fail} at Line~\ref{algline:fail} for some time point~$tp$. 
Because $tp.\#fp > 0$, at least one required task executes at time $tp.\mathit{time}$, so every feasible schedule must satisfy $\minC \leq f(tp.\mathit{time}) \leq \maxC$, where $f(tp.\mathit{time})$ denotes the resource usage.
Moreover, by Lemma~\ref{lem:profile-bounds}, $tp.\minP \leq f(tp.\mathit{time}) \leq tp.\maxP$ holds for any schedule consistent with the current domains.
Hence, any feasible schedule requires both $tp.\minP \leq \maxC$ and $tp.\maxP \geq \minC$. 
The condition at Line~\ref{algline:fail} ($tp.\minP > \maxC \lor tp.\maxP < \minC$) directly contradicts these requirements, proving that no feasible assignment exists and that returning \texttt{Fail} is sound.
\end{proof}

\subsection{Proof of Theorem~\ref{thm:init-complexity}}
\begin{proof}[Proof of Theorem~\ref{thm:init-complexity}]
At most $4n$ events are generated (two per task for the time window, plus two for the fixed part of each required task). Sorting them requires $\mathcal{O}(n \log n)$ time. All subsequent operations---time-point creation, accumulator updates, pointer assignments, and consistency checks---take $\mathcal{O}(1)$ per event, contributing $\mathcal{O}(n)$ in total. Hence the overall complexity is dominated by the sort.
\end{proof}

\subsection{Proof of Theorem~\ref{thm:tt-soundness}}
\begin{proof}[Proof of Theorem~\ref{thm:tt-soundness}]
The proof follows by checking each reduction rule against the exact profile bounds delivered by Lemma~\ref{lem:profile-bounds}. For any task~$i$ and any time point~$tp$ in the timeline, the values $tp.\minP$ and $tp.\maxP$ are the minimum and maximum resource usage compatible with the current domains, while $tp.\#fp$ counts the number of required fixed parts overlapping $tp$.
\begin{itemize}
  \item \emph{Forbid}: If $tp.\minP + \max(\minc_i,0) > \maxC$ at a time point in the forward scan, then every completion with $i$ starting at $tp.\mathit{time}$ would already exceed the upper capacity. Since $tp.\mathit{time} < \mine_i \leq e_i$, any schedule with $s_i \leq tp.\mathit{time}$ necessarily overlaps $tp$; thus $i$ cannot start at or before $tp.\mathit{time}$, which leaves $s_i \geq tp.\mathit{next}.\mathit{time}$ as the only valid domain reduction. The symmetric backward argument for $\maxe_i$ is identical.
  \item \emph{Mandatory (C3)}: Suppose $tp$ is outside the fixed part of $i$ ($tp.\mathit{time} \in [\mins_i,\maxe_i) \setminus \mathit{fp}_i$) and $tp.\#fp > 0$. If
  \[
      tp.\minP - \min(\minc_i,0) > \maxC \quad \text{or} \quad tp.\maxP - \max(\maxc_i,0) < \minC,
  \]
  then removing the contribution of $i$ from the profile at time~$tp$ would leave a resource load that is already incompatible with the capacity interval. Hence every feasible schedule must keep $i$ active at $tp$; otherwise the capacity constraint is violated. Furthermore, since all other tasks can collectively contribute at most $tp.\maxP - \max(\maxc_i, 0)$ at time~$tp$, task~$i$ must supply at least the remaining deficit $\minC - (tp.\maxP - \max(\maxc_i, 0))$, ensuring that tightening $\minc_i \geq \mathit{deficit}$ is sound (and symmetrically, $\maxc_i \leq \mathit{overload}$). This is the condition encoded by the rule \textit{checkIfMandatory}. Over the fixed part of a required task, the task is already known to execute, and the corresponding bound tightening is directly enforced by the \emph{Consumption} rule. For an optional task, the interval $[\maxs_i,\mine_i)$ is likewise checked by \textit{checkIfMandatory} in Step~3.
  \item \emph{Consumption}: If $\maxs_i < \mine_i$, then task~$i$ executes across this interval if it is executed. If $i \in R_{\text{init}}$, this interval is its fixed part $\mathit{fp}_i$, which contributed to the profile $\mathcal{L}$ during initialization, so its full fixed-part contribution is embedded in $\mathcal{L}$ and must be removed before checking feasibility: any value $c_i > \maxC - (tp.\minP - \minc_i)$ would force the minimum profile above $\maxC$; similarly, any value $c_i < \minC - (tp.\maxP - \maxc_i)$ would force the maximum profile below $\minC$. If $i \in O_{\text{init}}$, task~$i$ did not contribute its fixed part to $\mathcal{L}$ ($\mathit{fp}_i = \emptyset$); only its non-fixed contribution is removed (as in the (C3) check) to avoid counting the same task twice in the residual capacity. For tasks with $\maxs_i \geq \mine_i$ and $\mind_i \geq 1$, the task is guaranteed to execute at least at some point in $[\mine_i-1,\maxs_i]$, so the same argument applies to every time point in $[\mine_i-1,\maxs_i]$. So an assignment to $c_i$ is only invalid if it violates capacity across every point in the interval $[\mine_i-1,\maxs_i]$, which is exactly what the rule enforces. If $\mind_i = 0$, the task may execute with duration zero without consuming any resource, so its consumption is correctly not restricted.
  \item \emph{Duration}: If a sub-interval $[a,b)$ is excluded from $A$, then every schedule of duration~$b-a$ would force $i$ to overlap a point where the profile cannot satisfy the capacity range. Thus no feasible assignment can keep a duration longer than the maximum feasible span recorded by the algorithm.
\end{itemize}
Therefore every domain reduction performed by Algorithm~\ref{algo:timetabling} removes only values that cannot belong to a solution of $\gcumulative(T,\minC,\maxC)$.
\end{proof}

\subsection{Proof of Theorem~\ref{thm:tt-consistency}}
\begin{proof}[Proof of Theorem~\ref{thm:tt-consistency}]
The outer loop of Algorithm~\ref{algo:timetabling} (Line~\ref{algloop:timetabling}) visits every non-fixed task $i \in R_{\text{init}} \cup O_{\text{init}}$. For each task, the inner loops examine every relevant time point in the fixed timeline~$\mathcal{L}$, and each call to \textit{checkIfMandatory} is triggered on exactly the time points outside the fixed part that can witness a (C3)-conflict. We show that all conditions of Definition~\ref{def:tt-consistency} are satisfied upon termination:
\begin{enumerate}
  \item[\textbf{(C1)}] The forward loop at Line~\ref{algloop:ttfw} scans the timeline from $\mathit{tpAt}(\mins_i)$ and advances $\mins_i$ past all time points satisfying the (C1) condition. Upon exit, no (C1)-conflicting time point remains in $[\mins_i, \min(\maxs_i,\mine_i))$.
  \item[\textbf{(C2)}] The backward loop at Line~\ref{algloop:ttbw} scans the timeline from $\mathit{tpAt}(\maxe_i).\mathit{prev}$ and pulls $\maxe_i$ before every (C2)-conflicting point. Upon exit, no (C2)-conflicting time point remains in $[\max(\maxs_i,\mine_i),\maxe_i)$.
  \item[\textbf{(C3)}] The mandatory check is evaluated at every time point outside the fixed part visited in the forward, backward, and free-part scans (Lines~\ref{algline:fwmand},~\ref{algline:bwmand}, and~\ref{algline:freemand}), as well as across $[\maxs_i,\mine_i)$ for optional tasks (Line~\ref{algline:fpmand}). Whenever a (C3)-conflict arises, the necessary presence, time-window, and consumption conditions are enforced by \textit{checkIfMandatory}; hence no (C3)-conflicting point remains after termination.
  \item[\textbf{(C4)}] For tasks with $\maxs_i < \mine_i$, the loop at Line~\ref{algloop:ttfp} scans the interval $[\maxs_i,\mine_i)$ and distinguishes whether the task was initially required ($i \in R_{\text{init}}$) or optional ($i \in O_{\text{init}}$) when the timeline profile was built: when $i \in R_{\text{init}}$, its full contribution is removed at Line~\ref{algline:fph1}; when $i \in O_{\text{init}}$, only its non-fixed contribution is removed at Line~\ref{algline:fph2} to avoid double counting. The consumption domain is tightened until no (C4)-conflicting point remains.
\end{enumerate}
Together, (C1)--(C4) ensure pointwise consistency \textbf{(TC1)}. Furthermore:
\begin{itemize}
  \item \textbf{Free-part consumption consistency (TC2):} For tasks with $\maxs_i \geq \mine_i$ and $\mind_i \geq 1$, the loop at Line~\ref{algloop:ttfree} evaluates the minimum-overlapping interval $[\mine_i-1, \maxs_i]$ and updates the consumption bounds to $[\minc_i, \maxc_i]$ at Line~\ref{algline:h}, ensuring Condition~\textbf{(TC2)}.
  \item \textbf{Duration consistency (TC3):} For tasks with $\maxs_i \geq \mine_i$, the algorithm records the longest sub-interval of $[\mins_i,\maxe_i)$ that remains free of (C1)- and (C2)-conflicts; this is exactly the feasible duration-maximizing interval used to set~$\maxd_i$ at Line~\ref{algline:durationEnd}, satisfying Condition~\textbf{(TC3)}.
\end{itemize}
Since the timeline~$\mathcal{L}$ is fixed while these scans run, the state obtained when the algorithm terminates satisfies all conditions \textbf{(TC1)}--\textbf{(TC3)} of Definition~\ref{def:tt-consistency}. Hence Algorithm~\ref{algo:timetabling} is timetabling filtering consistent with respect to $\mathcal{L}$.
\end{proof}

\subsection{Proof of Theorem~\ref{thm:tt-complexity}}
\begin{proof}[Proof of Theorem~\ref{thm:tt-complexity}]
The timeline~$\mathcal{L}$ contains $\mathcal{O}(n)$ time points. The outer loop processes $\mathcal{O}(n)$ tasks. For each task, each inner loop scans at most $\mathcal{O}(n)$ time points with $\mathcal{O}(1)$ work per time point. The total cost is therefore $\mathcal{O}(n^2)$.
\end{proof}

\end{document}